%% file: paper.tex
\documentclass[]{bytedance_seed}

\usepackage[toc,page,header]{appendix}

\usepackage{minitoc}

\usepackage{booktabs}
\usepackage{threeparttable}
\usepackage{siunitx}
\usepackage[table]{xcolor}

\usepackage{makecell}

\usepackage{pifont}

\newcommand{\cmark}{\ding{51}}
\newcommand{\xmark}{\ding{55}}
\newcommand{\pmark}{\ensuremath{\triangle}}

\title{StartupBench: Benchmarking General-Purpose Agents on Market-Validated End-to-End Workflows}

\affiliation[1]{ByteDance Seed}
\affiliation[2]{Nanjing University}
\affiliation[3]{M-A-P}
\affiliation[4]{TokenWave.AI}

\contribution{Full author list in Contributions}

\abstract{
Recent advances in Large Language Models(LLMs) and agents have substantially improved the ability of AI systems to execute complex tasks. Yet existing benchmarks largely rely on researcher-selected tasks, leaving uncertain whether such progress extends to the work that real-world users actually demand from AI systems. We introduce \textbf{StartupBench}, an E2E agent benchmark grounded in market-validated AI startup products. Rather than defining tasks from pre-defined assumptions about useful agent capabilities, we systematically study AI products with demonstrated adoption, together with their product workflows and users, to identify real-world tasks for which AI has established practical demand across diverse professional domains. We translate these workflows into complete deliverable-oriented tasks and evaluate them with fine-grained rubrics capturing their complex requirements. Across representative models evaluated under a unified agent harness, even the strongest model successfully completes only approximately 30\% of StartupBench, despite making substantial partial progress on many tasks. Further analysis identifies aspects like complex instruction following and domain-specific expertise as major sources of failure. Our results reveal that many market-validated workflows remain beyond the reliable capabilities of current general-purpose agents, establishing StartupBench as an empirical measure of progress toward E2E completions of real-world user tasks.

}

\checkdata[Project Page]{\url{https://startupbench.github.io/}}

\date{\today}

\begin{document}
\maketitle


\input{sections/introduction}

\input{sections/relatedwork}
\input{sections/approach}
\input{sections/experiments}

\clearpage

\bibliographystyle{plainnat}
\bibliography{main}

\clearpage

\beginappendix

\input{sections/appendix}

\end{document}

%% file: sections/introduction.tex
\section{Introduction}

\begin{figure*}[t]
    \centering
A    \includegraphics[width=0.9\textwidth]{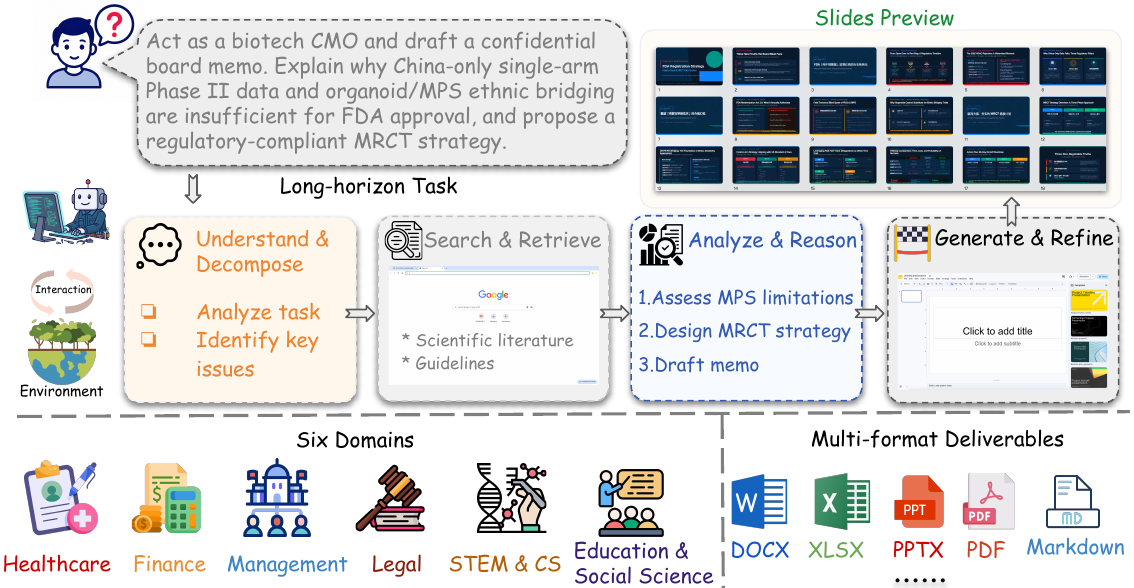}
    \caption{
    Overview of StartupBench: real AI-product workflows, long-horizon agent execution, six-domain coverage, and multi-format deliverable generation.
    }
    \label{fig:bench_overview}
\end{figure*}

Large language models (LLMs) are increasingly evolving from systems that primarily retrieve information or answer questions\cite{du2026supergpqa,mialon2024gaia,wang2024mmlu} into agents capable of completing complex real-world tasks. Current AI systems are expected to execute end-to-end (E2E) workflows, coordinate multiple capabilities, and produce deliverables that satisfy practical requirements under realistic constraints. As AI becomes increasingly integrated into professional work, autonomously completing user-delegated tasks has become an important measure of model capability. Evaluating whether AI systems can reliably translate their capabilities into professionally usable deliverables is therefore a central challenge for the next generation of benchmarks.

Recent benchmarks have made important progress toward evaluating E2E task execution under realistic settings. Nevertheless, several important limitations remain. First, many benchmark tasks are still largely defined from researchers' perspectives rather than grounded in workflows that have demonstrated practical demand through real-world AI adoption~\cite{phan2025humanity,patwardhan2025gdpval,mialon2024gaia}, limiting their ability to reflect authentic user needs and deliverable requirements. Second, although several benchmarks evaluate complete deliverables~\cite{tang2026workspace,sun2026agents}, their evaluation protocols are often substantially coarser than the deliverables themselves. Real-world work products typically involve numerous functional, structural, formatting, and domain-specific requirements that holistic evaluations cannot faithfully assess. Consequently, existing benchmarks still provide only limited evidence of \textit{how well current models and agents can satisfy complex real-world user requirements and reliably produce professionally usable deliverables}.

To address these limitations, we observe that AI-native startups offer a natural source of real-world AI workflows: their products have been validated through real-world adoption and commercial demand, providing direct evidence that users value and seek to delegate these tasks to AI. We therefore introduce \textbf{StartupBench}, a survey- and interview-driven benchmark that draws on these workflows to capture realistic E2E user requirements across domains. StartupBench spans 97 tasks across 6 primary domains and is built around the following design principles:

\begin{itemize}
\item \textbf{Adoption-grounded task sourcing.} We systematically survey AI-native startups and their products, complemented by product demonstrations and user interviews, to identify workflows with demonstrated real-world demand. We then construct these workflows into benchmark tasks that preserve their core user objectives and practical requirements.

\item \textbf{E2E deliverables.} Each task requires producing the complete deliverables that users ultimately expect in realistic workflows, rather than intermediate outputs or simplified demonstrations.

\item \textbf{Fine-grained rubric-based evaluation.} Each StartupBench task is evaluated using multiple independent rubrics covering distinct dimensions of the expected deliverable, including functionality, structure, formatting, and domain-specific quality, ensuring a comprehensive and faithful assessment of real-world task completion.

\end{itemize}

We evaluate representative models on StartupBench under  unified agentic harness. Results show that even the strongest model successfully completes only about 30\% of benchmark tasks, indicating that current general-purpose systems remain far from reliably producing professionally usable deliverables. Performance varies substantially across domains, with Finance, STEM \& Computer Science and Education \& Humanities proving particularly challenging. Deeper analysis further attributes these failures primarily to  limitations in complex instruction following and domain-specific expertise, highlighting promising directions for future foundation models and agent systems.

To summarize, our contributions are as follows:

\begin{itemize}
    \item We propose a survey- and interview-driven methodology for constructing agent benchmarks from real AI adoption scenarios.
    \item We construct StartupBench, a benchmark built through AI-native startup research, enterprise-user interviews, user-oriented task abstraction, expert data construction, and multi-stage quality control.
    \item We present a comprehensive empirical study of current foundation models and general-purpose agents on startup-derived workflows, providing a practical evaluation signal for assessing whether AI agents are progressing from answering questions toward reliably completing real work.
\end{itemize}


%% file: sections/relatedwork.tex
\section{Related Work}
\subsection{Commercial AI agents and Vertical Workflows}
Recent advances in foundation models have enabled AI agents to gradually acquire planning, tool-use, file-processing, and multi-step execution capabilities \citep{yao2023react, schick2023toolformer}, pushing AI systems from general conversational assistants toward workflow-oriented products. In practice, commercial AI products increasingly package these capabilities into vertical workflows such as document processing, data analysis, financial research, enterprise operations, and knowledge work. This trend suggests that real-world AI value is not determined only by isolated model capabilities, but also by whether these capabilities can be productized into useful work products. However, existing academic benchmarks rarely use commercial adoption itself as a source of task discovery. StartupBench addresses this gap by deriving tasks from market-validated AI-native startups and their deep users, moving task construction from researcher-defined settings toward demand-driven workflow construction.

\subsection{End-to-end evaluation of AI agents}
Agent benchmarks have evolved from capability evaluation toward end-to-end task completion. Early benchmarks mainly focus on tool use, multi-step reasoning, and general assistant abilities, as in AgentBench and GAIA \citep{liu2024agentbench, mialon2024gaia}. Later work extends evaluation to more realistic interactive environments, including web interaction and computer-use tasks \citep{zhou2024webarena, xieosworld}. More recent benchmarks further move toward professional work and economically meaningful tasks, covering enterprise software, simulated workplaces, occupational tasks, and long-horizon workflows \citep{drouin2403workarena, xu2026theagentcompany, patwardhan2025gdpval, hu2026occubenchevaluatingaiagents, zhu2026workflowgymlonghorizonevaluationcomputeruse,meng2026dvworld}. These works reflect a clear transition from isolated capability tests to realistic task-completion evaluation. StartupBench follows this direction, but differs in its task source: rather than relying primarily on public environments, occupational taxonomies, or simulated workplaces, it derives tasks from AI-native startup products, ToB user needs, and expected business deliverables.
\subsection{Agent-as-a-Judge for complex deliverable evaluation}
As agent tasks shift from short answers to heterogeneous deliverables such as reports, spreadsheets, slides, code patches, and structured files, evaluation must move from capability-centric scoring to deliverable-centric assessment\cite{you2026agent}. LLM-as-a-Judge has been widely used for open-ended response evaluation, with MT-Bench and Prometheus studying preference-based or rubric-conditioned judging \citep{zheng2023judging, kim2024prometheus}. For agentic workflows, however, judging often requires evidence retrieval, file inspection, state verification, and constraint checking over final artifacts. Agent-as-a-Judge extends LLM judging by equipping evaluators with tools and environment interaction, while AJ-Bench further studies judge agents on information acquisition, state verification, and process verification \citep{zhuge2024agent, shi2026aj}. StartupBench adopts this deliverable-centric evaluation paradigm by combining expert rubrics with automated judge agents that inspect original artifacts and evidence views, evaluating whether agents can transform model capabilities into usable work products.


%% file: sections/approach.tex
\section{StartupBench}


\begin{table}[t]
\caption{Comparison of StartupBench with existing benchmarks. E2E denotes end-to-end workflow completion; Item-wise Eval. denotes criterion-level judging. \cmark, \xmark, and \pmark indicate full, no, and partial support.}
\label{tab:main_stat}
\centering
\resizebox{\textwidth}{!}{
\begin{tabular}{lccccccc}
\toprule
\textbf{Benchmark} &
\textbf{\thead{Task Source}} &
\textbf{\thead{Final \\ Output}} &
\textbf{E2E} &
\textbf{\thead{Item-wise \\ Eval.}} &
\textbf{\thead{Process \\ Logic}} &
\textbf{\thead{Open-\\ended}} &
\textbf{\thead{Evaluation \\ Method}} \\
\midrule

GAIA \citep{mialon2024gaia}
& Manual
& Text
& \pmark
& \xmark
& \xmark
& \xmark
& Rule \\

OSWorld \citep{xieosworld}
& Real+Manual
& Env State
& \pmark
& \xmark
& \xmark
& \xmark
& Execution \\

DAComp \citep{lei2025dacomp}
& Real+Manual
& Workspace
& \cmark
& \xmark
& \xmark
& \cmark
& Hybrid \\

GDPval \citep{patwardhan2025gdpval}
& Manual
& Work products
& \cmark
& \xmark
& \xmark
& \pmark
& Manual \\

Workspace-Bench \citep{tang2026workspace}
& Sim.+Manual
& Workspace 
& \cmark
& \xmark
& \xmark
& \cmark
& Agent-as-Judge \\

OneMillionBench \citep{yang2026onemillion}
& Expert-authored
& Text
& \pmark
& \xmark
& \pmark
& \cmark
& LLM-judge \\

Agents' Last Exam \citep{sun2026agents}
& Occupation-guided
& Workspace
& \cmark
& \xmark
& \pmark
& \pmark
&  Hybrid \\

OfficeQA Pro \citep{opsahl2026officeqa}
& Corpus-grounded
& Text
& \xmark
& \xmark
& \xmark
& \xmark
& Answer match \\

\midrule

\textbf{StartupBench (Ours)}
& \textbf{\thead{Market-vetted\\+ Expert-built}}
& \textbf{Workspace}
& \cmark
& \cmark
& \cmark
& \cmark
& \textbf{Agent-as-Judge} \\

\bottomrule
\end{tabular}
}
\end{table}

\begin{figure*}[t]
    \centering
    \includegraphics[width=0.99\textwidth]{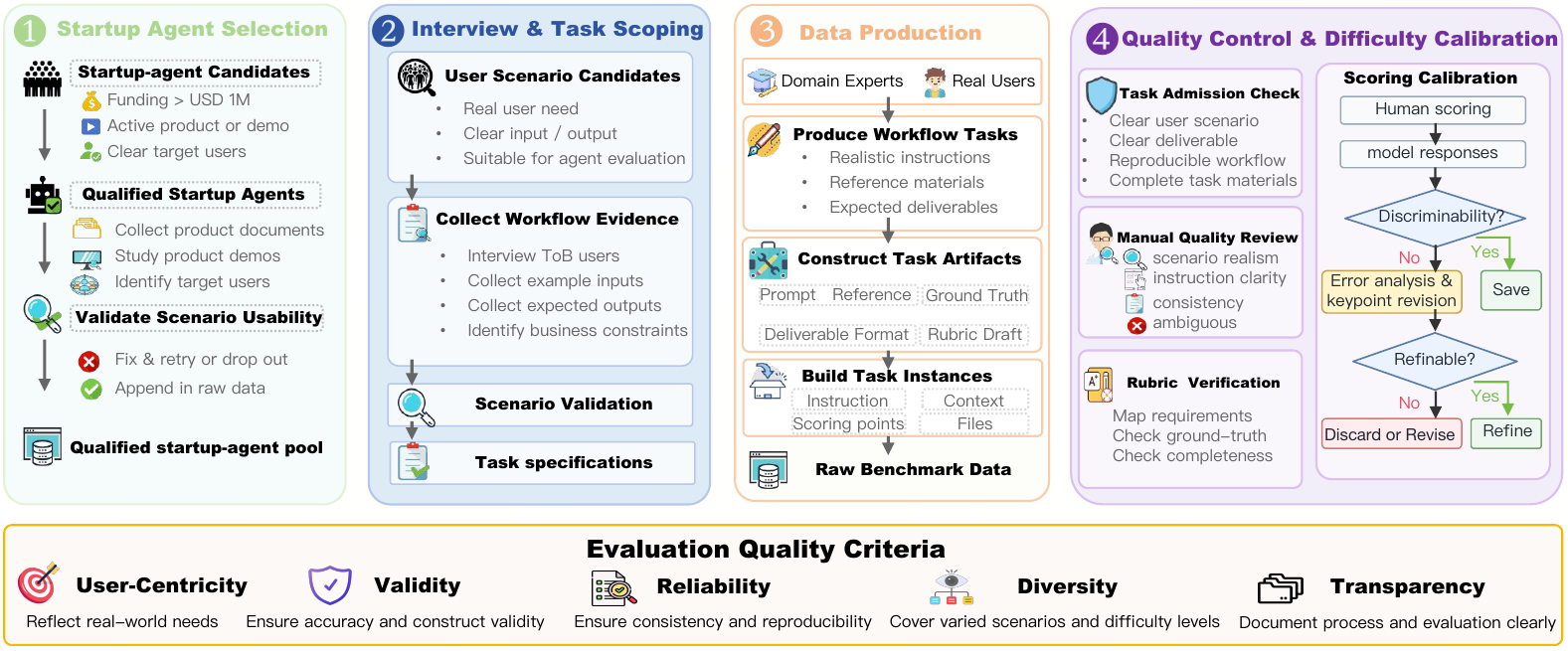}
    \caption{
    StartupBench data construction pipeline. We select market-validated startup agents, collect real user workflows, construct task artifacts and instances, and calibrate quality and difficulty through multi-stage review.
    }
    \label{fig:data_workflow}
\end{figure*}

StartupBench is designed to evaluate whether models and agents can complete realistic workflows that users already delegate to AI products. This objective requires data that reflects authentic user demand while supporting controlled evaluation. We therefore require each task to be realistic, answerable, evaluable, and discriminative: it should originate from actual AI product usage, provide sufficient information for a valid solution, specify clear output requirements and assessment criteria, and remain challenging enough to distinguish current models and agents.

Neither manual task design nor direct extraction from product demonstrations is sufficient to obtain such data. The former is controllable but may be detached from real demand, while the latter is realistic but often lacks context, success criteria, or reproducible specifications. We therefore adopt a survey- and interview-driven multi-stage pipeline shown in Figure~\ref{fig:data_workflow}.  We first survey market-validated AI-native startups, interview deep users to identify usage contexts, task goals, and expected deliverables, recruit domain experts to convert validated scenarios into benchmark instances, and finally apply quality control for realism, answerability, evaluability, and discriminative difficulty.
\subsection{Dataset Construction}

\subsubsection{Startup Survey for Market-Validated Agent Scenarios}

We first survey AI-native startups that build agent products across diverse domains. To identify workflows with market validation, we retain startups with more than USD 1M in funding and require evidence of real adoption, either through paid usage or substantial user traction. Funding provides a coarse signal of market expectation, while payment or user-scale evidence indicates that the product is used beyond exploratory demonstrations. The output of this stage is a pool of candidate products and workflows for user interviews. During this stage we collect 20+ start-up agents as the basis for subsequent stage.

\subsubsection{User Interviews for Scenario and Requirement Discovery}

For each candidate Startup-product, we interview deep users of the corresponding agent from different domains to recover practical usage beyond public product descriptions. The interviews elicit the usage context, user goal, input information, expected deliverable, success criteria, and common constraints. We summarize these findings into scenario specifications and representative demo cases, which serve as anchors for later annotation stage.
During this stage we interview over 30 deep users and collected at least one demo case for each candidate Agent.

\subsubsection{Expert Construction of Domain Tasks}

Given the interview-derived workflow specifications and representative user scenarios as a reference or seed task, we recruit over 50 domain experts to reconstruct benchmark tasks that preserve the original workflow objectives, practical constraints, and expected deliverables while adapting them into reproducible evaluation instances. Rather than designing synthetic problems, experts standardize authentic user requests into benchmark tasks that faithfully represent real professional workflows. 

Every task is required to satisfy three principles: it must originate from a realistic work request that users would naturally delegate to AI agents, involve an end-to-end workflow rather than an isolated subtask, and produce concrete professional deliverables with clearly defined success criteria and fine-grained evaluation rubrics for objective evaluation.

To standardize these tasks for evaluation, each task is represented as a triple
\[
\mathcal{T}=(q,\mathcal{E},\mathcal{R}),
\]
where $q$ is a natural-language user request describing the task, $\mathcal{E}$ denotes the workspace containing all input files and resources required to complete the task, and $\mathcal{R}=\{(p_i,w_i)\}_{i=1}^{n}$ is a set of weighted evaluation rubrics that assess complementary aspects of deliverable quality. Detailed task specifications and annotation guidelines are provided in Appendix~\ref{app:task_construction}. During this stage, we collect over 150 candidate tasks for further quality control.

\subsubsection{Quality Control}
To ensure benchmark quality and consistency, every task undergoes the following multi-stage quality control process:

\textbf{Expert Cross Validation.}
Every task is independently reviewed by at least one additional domain expert.
Reviewers verify four aspects of task quality:
(1) the authenticity and correctness of the task description and workspace,
(2) the fidelity of the reconstructed workflow,
(3) the validity of the evaluation rubrics, and
(4) the consistency between the reference deliverable and the finalized evaluation protocol.
Detailed review criteria are provided in Appendix~\ref{app:cross_valid}.

\textbf{Difficulty Control.}We calibrate task difficulty through pilot executions using several frontier models under the same evaluation harness adopted in the benchmark\footnote{We randomly select models from GPT-5.5, Seed-2.1-Pro and GLM-5.1 as the model for pilot executions, intending to construct a challenging and discriminative benchmark rather than to estimate performance over the natural task distribution.}. Tasks that are consistently completed with high quality are excluded due to limited discriminative value. Pilot results also serve as an additional quality assurance stage: when failures are attributed to ambiguous instructions, incomplete workspace information, or inadequately specified evaluation criteria rather than intrinsic task difficulty, the corresponding task is revised and re-validated before inclusion.

\subsection{Statistics for StartupBench}

\textbf{StartupBench} contains 97 real-world workflow tasks across 6 top-level domains: Medical \& HealthCare, Finance, Legal, Business \& Management, STEM \& Computer Science, and Education \& Humanities. These tasks are designed to reflect deliverable-oriented workflows in realistic office settings, requiring agents to understand task contexts, follow complex constraints, process heterogeneous files, and produce usable work products. For evaluation, each task is evaluated using an average of \textbf{25.3} fine-grained rubrics, spanning \textbf{6 dimensions and 3 importance levels} , enabling detailed assessment of complex, long-horizon workflows and diverse work deliverables. The  summary of the domain coverage and main output formats of StartupBench is shown in Table~\ref{tab:benchmark_statistics}.

\begin{table}[t]
\centering
\small
\setlength{\tabcolsep}{12pt}
\renewcommand{\arraystretch}{1.12}
\begin{threeparttable}
\caption{Statistics of StartupBench. The benchmark contains 97 real-world workflow tasks across six top-level domains and requires diverse deliverable formats.}
\label{tab:benchmark_statistics}

\begin{tabular}{l S[table-format=3.0] S[table-format=3.1]}
\toprule
\textbf{Category} & {\textbf{Count}} & {\textbf{Percentage (\%)}} \\
\midrule

\rowcolor{gray!8}
\multicolumn{3}{l}{\textit{Overall statistics}} \\
Total tasks & 97 & 100.0 \\
Top-level domains & 6 & \multicolumn{1}{c}{--} \\

\midrule
\rowcolor{gray!8}
\multicolumn{3}{l}{\textit{Domain distribution}} \\
Medical \& HealthCare & 21 & 21.6 \\
Finance & 18 & 18.6 \\
Legal & 16 & 16.5 \\
Business \& Management & 19 & 19.6 \\
STEM \& Computer Science & 16 & 16.5 \\
Education \& Humanities & 7 & 7.2 \\

\bottomrule
\end{tabular}

\begin{tablenotes}[flushleft]
\footnotesize
\item \textit{Output formats:} DOCX, XLSX, PPTX, PDF, Markdown, images, and text-based deliverables.
\end{tablenotes}
\end{threeparttable}
\end{table}

\subsection{Automatic Evaluation of StartupBench Tasks}

As stated above, every task in StartupBench is defined as a triple $\mathcal{T}=(q,\mathcal{E},\mathcal{R})$. Given $(q,\mathcal{E})$, the evaluated model produces a set of deliverables $\mathcal{D}$. Before evaluation, we construct a lightweight evidence view $\mathcal{V}(\mathcal{D})$, consisting of extracted textual content and rendered page images, to facilitate navigation over heterogeneous artifacts while allowing the judge to access the original files whenever necessary.

Rather than relying on a single holistic judgment, StartupBench evaluates each rubric item independently through a dedicated AgentJudge session. Most tasks contain 20+ fine-grained rubric items covering diverse aspects of the expected deliverables, including functionality, completeness, formatting, and domain-specific requirements. Evaluating these criteria separately allows each rubric to receive dedicated reasoning and tool usage without interference from unrelated evaluation dimensions, while ensuring that every requirement is assessed explicitly. The final task score is then obtained by aggregating the outcomes of all rubric-level evaluations.

Following \textbf{Agent-as-a-Judge}, each rubric evaluation is carried out by a lightweight agent rather than a single forward pass. Let $J$ denote the judging prompt and evaluation protocol used by the AgentJudge, given the task specification, deliverables, evidence view, and the target rubric item $p_i$, the judge produces both a binary decision $y_i\in\{0,1\}$ indicating whether the rubric is satisfied and a textual justification $e_i$:

\begin{equation}
(y_i,e_i) = \mathrm{AgentJudge}\big(J, q,
\mathcal{D}, \mathcal{V}(\mathcal{D}), p_i\big),
\label{eq}
\end{equation}

The final task score is computed by aggregating the weighted decisions across all rubric items:

\begin{equation}
\mathrm{score}(\mathcal{D},\mathcal{T})
= \frac{\sum_{i=1}^{n} w_i y_i}{\sum_{i=1}^{n} w_i}
\in [0,1].
\label{eq2}
\end{equation}

%% file: sections/experiments.tex

\section{Experiments}
\label{sec:experiments}

\subsection{Experimental Setup}

\label{sec:exp-setup}

We evaluate 9 representative models on StartupBench, spanning both closed- and open-source families:
\begin{itemize}
    \item  \textbf{Closed-source}:  GPT-5.6-sol~\cite{openai2026gpt56}, GPT-5.5 ~\cite{openai_gpt55}, Gemini-3.1-Pro~\cite{GoogleCloud2026Gemini3.1Pro}, Seed-2.1-Pro~\cite{bytedance2026seed21},Qwen-3.6-Max~\cite{qwen36_max_preview}
    
    \item \textbf{Open-source}: DeepSeek-V4-Pro~\cite{deepseek2026v4},Kimi-K3 \cite{team2026kimi}, Kimi-K2.6~\cite{moonshot2026k2_6},  GLM-5.1~\cite{zeng2026glm}. 
\end{itemize}
To improve the reliability of our results, we conduct \textbf{3} independent runs for each model and report 95\% bootstrap confidence intervals for the task scores, computed using 10{,}000 resamples.
All models are executed under the same Nanobot harness with an identical tool configuration, and every task is capped at a maximum of \textbf{200} interaction steps. 
For each task, the agent is initialized with the task-specific workspace and input query, and its deliverables are collected from the designated output directory. Following the evaluation protocol described in the previous section, the deliverables are evaluated by the judge agent against the task specification, workspace context, and rubric set.

For automatic evaluation, the judge agent also runs under the same Nanobot harness to ensure a consistent execution environment, while using GPT-5.5 as the underlying judge model. Apart from the continuous task score, we also report \textbf{success rate}, the fraction of runs that are successfully completed, pooled over the three runs. A task is considered successful if its final score is at least \textbf{90}; otherwise, it is counted as a failure.

\subsection{Main Results}
\label{sec:exp-overall}

\begin{figure}[t]
  \centering
  \includegraphics[width=0.95\linewidth]{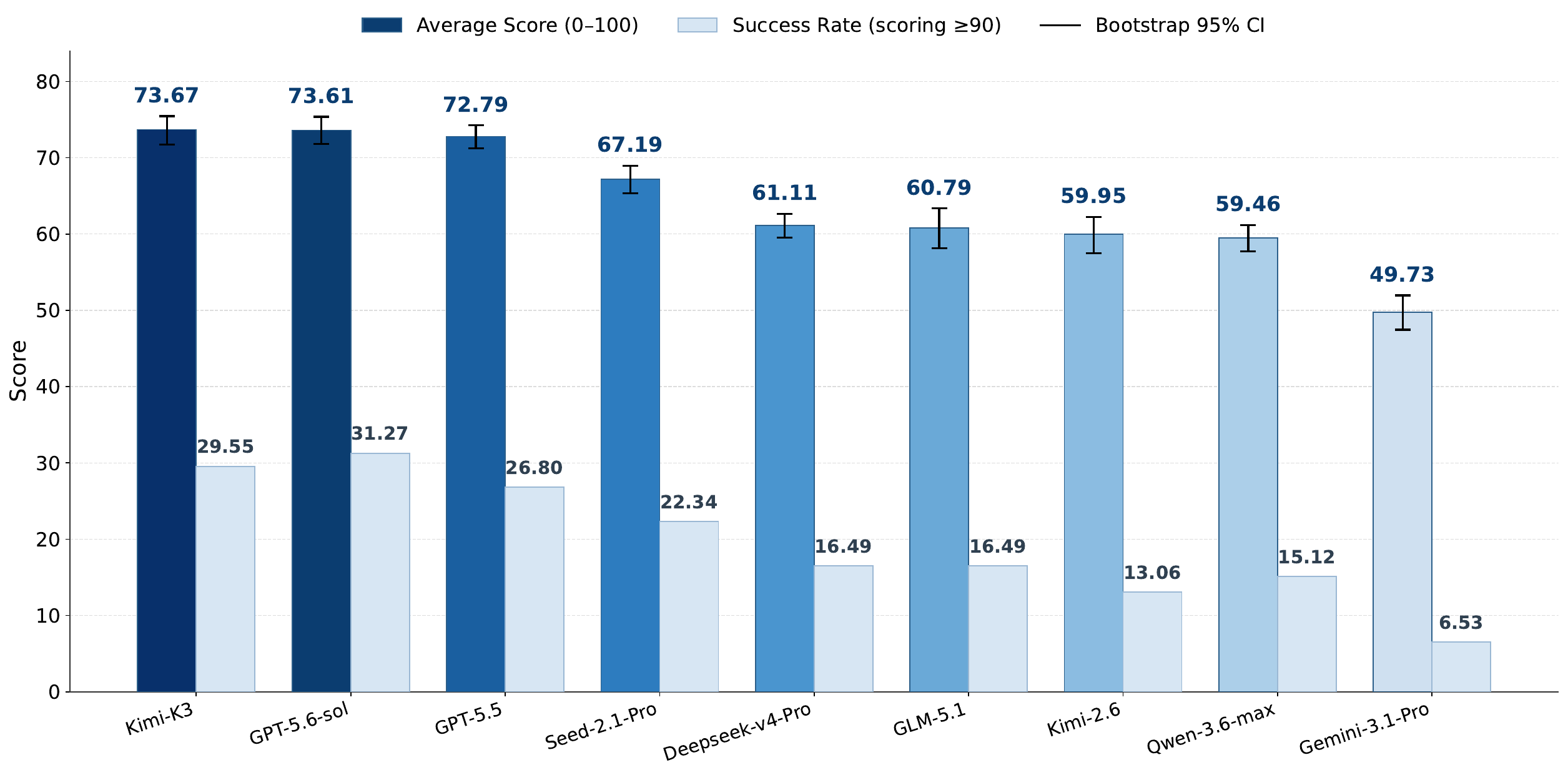}
  \caption{Leaderboard of StartupBench.}
  \label{fig:overall-leaderboard}
\end{figure}

\begin{table*}[t]
  \centering
  \small
  \setlength{\tabcolsep}{5pt}
  \renewcommand{\arraystretch}{1.15}
  \caption{Model performance of StartupBench, averaged over three independent runs.
    \textbf{Top block:} importance-weighted mean per-task score ($0$--$100$);
    \textbf{bottom block:} success rate, the fraction of samples scoring $\ge 90$ (\%),
    pooled over the three runs. Per-column \textbf{best in bold},
    \underline{second-best underlined}.}
  \label{tab:leaderboard}
  \begin{tabular}{lccccccc}
    \toprule
    \textbf{Model} & \textbf{Overall}
      & \textbf{Med} & \textbf{Finance} & \textbf{Legal}
      & \textbf{Business} & \textbf{STEM} & \textbf{Education} \\

    \midrule
    \multicolumn{8}{c}{\textbf{Average score}} \\
    \midrule
    Kimi-K3 & \textbf{73.67} & \textbf{80.03} & \textbf{62.38} & 69.45 & \textbf{83.90} & 68.34 & \textbf{77.71} \\
    
    GPT-5.6-sol & \underline{73.61} & \underline{79.32} & 61.66 & \textbf{73.53} & 77.61 & \textbf{74.77} & \underline{73.94} \\ 
    
    GPT-5.5         & 72.79 & 78.59 & \underline{62.09} & \underline{71.75} & \underline{79.59} & \underline{72.14} & 68.32 \\
    Seed-2.1-Pro    & 67.19 & 73.15 & 57.26 & 61.35 & 77.83 & 65.61 & 62.91 \\
    Kimi-K2.6       & 59.95 & 67.70 & 51.17 & 54.62 & 71.91 & 51.38 & 58.60 \\
    GLM-5.1         & 60.79 & 65.66 & 51.22 & 50.20 & 78.53 & 52.18 & 66.50 \\
    Qwen-3.6-Max    & 59.46 & 66.51 & 52.45 & 50.33 & 75.03 & 48.83 & 59.26 \\
    DeepSeek-V4-Pro & 61.11 & 69.42 & 51.11 & 54.53 & 73.69 & 54.89 & 57.03 \\
    Gemini-3.1-Pro  & 49.73 & 54.05 & 41.00 & 43.49 & 61.18 & 45.88 & 51.23 \\
    \midrule
    \multicolumn{8}{c}{\textbf{Success rate}\, ($\ge90$)} \\
    \midrule
    Kimi-K3 & \underline{29.55} & \textbf{34.92} & 20.37 & 20.83 & \textbf{52.63} & 16.67 & \underline{23.81} \\
    GPT-5.6-sol & \textbf{31.27} & \underline{33.33} & \textbf{27.78} & \underline{22.92} & 38.60 & \textbf{33.33} & \textbf{28.57} \\
    
    GPT-5.5         & 26.80 & 30.16 & \underline{22.22} & \textbf{25.00} & 38.60 & \underline{22.92} & 9.52 \\
    Seed-2.1-Pro    & 22.34 & 19.05 & 16.67 & 14.58 & \underline{45.61} & 20.83 & 4.76 \\
    Kimi-K2.6       & 13.06 & 15.87 & 14.81 & 6.25 & 24.56 & 4.17 & 4.76 \\
    GLM-5.1         & 16.49 & 12.70 & 16.67 & 8.33 & 36.84 & 8.33 & 9.52 \\
    Qwen-3.6-Max    & 15.12 & 11.11 & 16.67 & 10.42 & 33.33 & 6.25 & 4.76 \\
    DeepSeek-V4-Pro & 16.49 & 20.63 & 16.67 & 8.33 & 29.82 & 10.42 & 0.00 \\
    Gemini-3.1-Pro  & 6.53 & 6.35 & 11.11 & 6.25 & 8.77 & 0.00 & 4.76 \\
    \bottomrule
  \end{tabular}
\end{table*}

\textbf{Startup tasks remain far from saturated under a general-purpose agent harness.} Although the strongest models, Kimi-K3 and GPT-5.6-sol, achieve average scores of 73.67\% and 73.61\%, respectively, no model successfully completes even one third of the benchmark under the strict acceptance criterion (score $\ge 90$). More broadly, the average scores of most models lie between roughly 55 and 75, indicating that current agents are generally capable of making substantial progress toward solving realistic professional workflows. However, high average scores do not necessarily translate into high task completion rates. This is particularly evident for the Kimi series: Kimi-K3 achieves the highest average score overall but a lower success rate than GPT-5.6-sol, while Kimi-K2.6 similarly attains a higher average score than Qwen-3.6-Max but a lower success rate. These discrepancies suggest that some models can satisfy a substantial fraction of task requirements while still falling short of complete delivery. More generally, all models exhibit substantially lower success rate than average score, suggesting that the primary bottleneck is no longer executing large portions of a workflow, but consistently producing artifacts that satisfy the standards required for direct professional use.

This gap has two important implications. First, it helps explain why specialized vertical agents continue to provide practical value in many market-validated workflows, where consistently meeting professional standards remains essential. Second, it suggests that the next frontier for foundation models is not merely performing professional tasks, but reliably producing work products that users can adopt without further verification or revision. We further investigate the underlying causes of this gap in Section~\ref{sec:exp-failures}.

\textbf{Performance varies substantially across professional domains.} Table 3 reveals that current general-purpose agents do not exhibit uniform competence across real-world workflows. Among the six domains, Business is consistently the easiest: every model achieves an average score above 60, and the average success rate across models is also substantially higher than in any other domain. In contrast, Finance is the most challenging, with the lowest average score of 54.48\%. The large gap between Business and Finance suggests that current models handle structured, relatively standardized tasks more reliably than workflows requiring sustained quantitative reasoning, cross-document consistency, and multi-step verification.

The results also reveal pronounced domain specialization. While Kimi-K3 and GPT-5.6-sol achieve nearly identical average scores overall (73.67 vs. 73.61), their strengths differ substantially across domains. Under success rate, Kimi-K3 performs best in Medical and Business, whereas GPT-5.6-sol leads in Finance, STEM, and Education; GPT-5.5 remains strongest in Legal. A similar pattern emerges in average score, where Kimi-K3 leads in Medical, Business, Finance and Education, GPT-5.6-sol in Legal and STEM. Thus, even among the strongest models, no single agent consistently dominates across professional domains. These results suggest that current general-purpose agents still exhibit substantial domain-dependent variation, leaving considerable room for improving the robustness and transferability of end-to-end task execution across heterogeneous professional workflows.


\subsection{Effectiveness of Proposed Evaluation Framework}
\label{sec:exp-agreement}


To validate the reliability of our evaluation framework, we compare its judgments against expert annotations. We sample outputs from two representative models for all tasks and ask domain experts to independently evaluate each submission using the predefined rubrics. At the rubric level, the automatic evaluator achieves an overall agreement of \textbf{92.78\%} with expert judgments. We further examine whether these local agreements translate into consistent task-level decisions by comparing the resulting success labels, where a submission is considered successful if its aggregated score reaches 90. The automatic and expert evaluations agree on \textbf{92.84\%} of these success decisions. Notably, the similarly high agreement at both rubric and success levels, despite their substantially different label distributions, suggests that the observed consistency is not merely an artifact of label imbalance.

We further investigate the necessity of evaluating each rubric independently. As an ablation, instead of assigning one rubric to one judge agent, we directly provide the complete model output together with the full rubric list to a single judge agent, requiring it to score all rubrics in one end-to-end inference. This alternative reduces the agreement with human experts to \textit{83\%}. More importantly, the holistic setting proves considerably less stable in practice. Since the judge agent must simultaneously perform long-context reasoning, maintain consistency across numerous evaluation criteria, and finally generate a structured output covering all rubric scores, failures such as malformed outputs or incomplete formatting frequently occur, forcing repeated executions. On average, each evaluation requires more than two runs before a valid result is obtained. In contrast, our rubric-wise evaluation decomposes the assessment into a collection of independent, lightweight judging tasks, yielding both substantially higher agreement with human experts and significantly better execution robustness.

\subsection{Failure-Mode Analysis}

\subsubsection{Outcome-level Failure Analysis}
\label{sec:exp-format}

\begin{figure*}[t]
    \centering
    \includegraphics[width=0.99\textwidth]{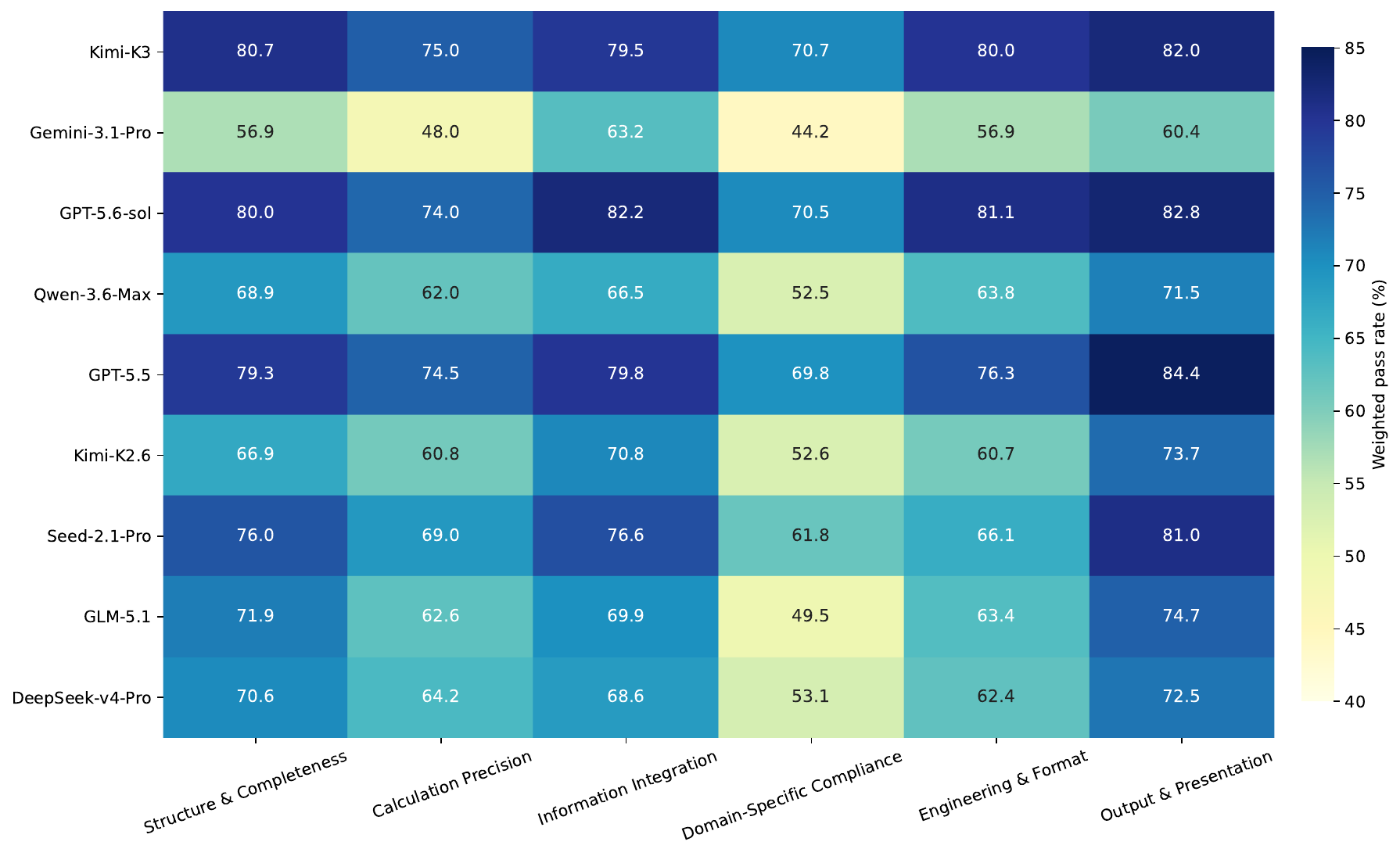}
    \caption{
Performance across rubric dimensions. Importance-weighted pass rates (\%) of different models on the six rubric dimensions in StartupBench. Higher values indicate better performance under the corresponding evaluation dimension.
    }
    \label{fig:dimension-rubric}
\end{figure*}

\textbf{Capability differences across rubric dimensions.} Figure~\ref{fig:dimension-rubric} further breaks down model performance by rubric dimension. Overall, current models achieve consistently higher scores on Structure \& Completeness, Information Integration, and especially Output \& Presentation, indicating that they are generally capable of organizing heterogeneous information into well-structured and visually polished deliverables. In contrast, Domain-Specific Compliance emerges as the most challenging dimension across all evaluated models, while Calculation Precision also remains substantially weaker than the other dimensions. This suggests that although modern foundation models can produce artifacts that appear complete and professional, reliably satisfying domain-specific regulations, business conventions, and numerical correctness continues to be a major obstacle for real-world deployment.

The two strongest models, GPT-5.6-sol and Kimi-K3, outperform the remaining systems across nearly all rubric dimensions rather than relying on a single specialized capability. Their largest advantages are observed in Structure \& Completeness, Calculation Precision, Information Integration, and Output \& Presentation, demonstrating stronger end-to-end workflow execution and artifact construction capabilities. Nevertheless, even these leading models obtain their lowest scores on Domain-Specific Compliance, indicating that adherence to professional standards and domain-specific requirements remains the primary bottleneck even for state-of-the-art systems.

\textbf{Models often satisfy peripheral requirements while missing what matters most.}
Table~\ref{tab:corefail} reports task-normalized rubric satisfaction rates across the three importance levels. Averaged across models, satisfaction decreases from \textbf{68.67\%} for Auxiliary rubrics to \textbf{65.89\%} for Important rubrics and \textbf{63.45\%} for Core rubrics, with 8 of the 9 models exhibiting the same overall trend. Importantly, these levels reflect each requirement's contribution to successful task completion rather than its expected difficulty. This pattern reveals a characteristic failure mode of current agents: they can often fulfill auxiliary or surface-level requirements while failing on the core requirements that determine whether the task is actually completed successfully. In other words, substantial apparent progress can be driven by satisfying less consequential parts of the task, while critical omissions still prevent the final deliverable from being practically usable.

\begin{figure*}[t]
    \centering
    \includegraphics[width=0.8\textwidth]{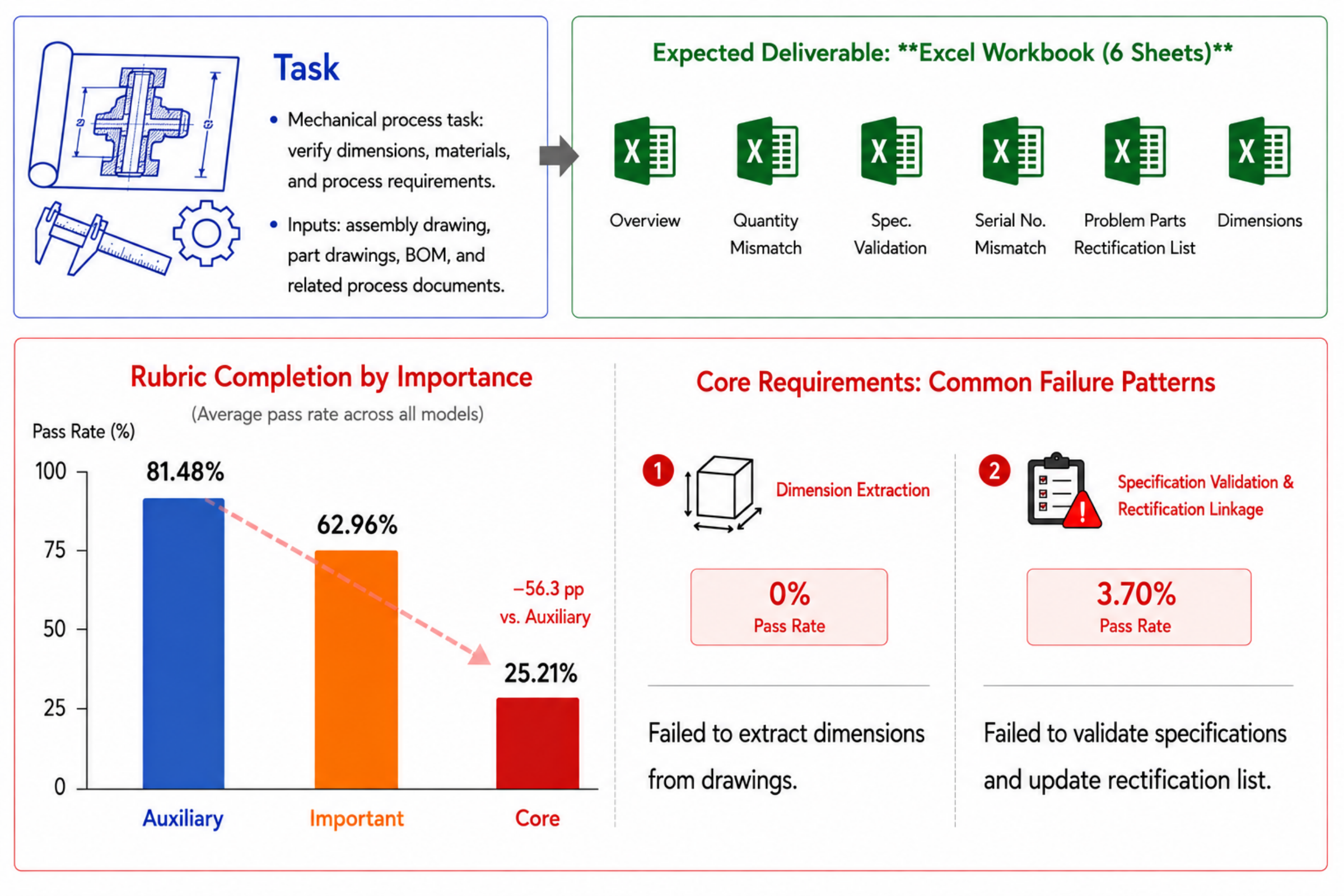} 
    \caption{
A representative case of models failing on core rubrics though passing auxiliary rubrics on StartupBench.
    }
    \label{fig:corefail}
\end{figure*}

A typical case of models struggling with requirements central to task completion is shown in Figure~\ref{fig:corefail}. In this mechanical process task, models often complete peripheral workbook requirements but fail on the core technical operations: extracting dimensions accurately from engineering drawings, validating specifications against the provided documents, and linking identified discrepancies to the corresponding rectification actions. These failures prevent the resulting workbook from fulfilling its primary purpose even when many auxiliary requirements are successfully completed.

\begin{table}[t]
\centering
\small
\caption{Task-normalized rubric satisfaction rates (\%) across different importance levels. 
For each model--task--trial, we first compute the satisfaction rate within each importance level and then average across tasks and trials.}
\label{tab:corefail}
\begin{tabular}{lccc}
\toprule
\textbf{Model} 
& \textbf{Auxiliary }
& \textbf{Important }
& \textbf{Core } \\
\midrule
Kimi-K3          & 77.89 & 75.12 & 73.40 \\
GPT-5.6-sol      & 77.59 & 74.47 & 73.07 \\
GPT-5.5          & 75.12 & 75.09 & 72.07 \\
Seed-2.1-Pro         & 70.83 & 69.85 & 65.77 \\
Kimi-K2.6        & 66.64 & 63.56 & 57.82 \\
GLM-5.1          & 64.58 & 61.40 & 60.68 \\
Qwen-3.6-Max     & 65.39 & 62.17 & 58.08 \\
DeepSeek-V4-Pro  & 66.50 & 62.97 & 60.05 \\
Gemini-3.1-Pro       & 53.48 & 48.41 & 50.13 \\
\midrule
\textbf{Average} & \textbf{68.67} & \textbf{65.89} & \textbf{63.45} \\
\bottomrule
\end{tabular}
\end{table}

\textbf{Output format non-compliance.}
Some failures arise from models not producing deliverables in the file type explicitly required by the task. Unlike most failure modes discussed above, this requirement involves neither complex reasoning nor domain knowledge and can be verified deterministically from file extensions. Nevertheless, no model achieves perfect compliance on the 56 tasks with explicit output-format requirements (Table~\ref{tab:format-compliance}), with success rates ranging from $97.6\%$ to $86.3\%$. Most violations fall into two recurring patterns: generating an incorrect file type (e.g., returning \texttt{.md}/\texttt{.txt} instead of the required \texttt{.pdf}) and omitting part of the requested deliverables (e.g., producing only one of the required \texttt{.xlsx} and \texttt{.docx} files). These errors suggest that even elementary deliverable-level constraints remain a non-negligible source of failures for current foundation models.

\begin{table}[t]
\centering
\footnotesize
\caption{Output-format compliance on the 56 StartupBench tasks with explicit output-file requirement rubrics. Models are ranked by compliance rate.}
\label{tab:format-compliance}
\begin{tabular}{lc}
\toprule
\textbf{Model} & \textbf{Compliance (\%)}\\
\midrule
Kimi-K3           & 97.6 \\
Seed-2.1-Pro      & 95.2 \\
GPT-5.6-sol       & 94.6 \\
Qwen-3.6-Max      & 94.6 \\
DeepSeek-V4-Pro   & 94.0 \\
GPT-5.5           & 93.5 \\
Kimi-K2.6         & 92.9 \\
Gemini-3.1-Pro    & 86.9 \\
GLM-5.1           & 86.3 \\
\bottomrule
\end{tabular}
\end{table}

\subsubsection{Behavior-level Failure Analysis}
\label{sec:exp-failures}
\textbf{Complex Instruction Following.} Complex real-world workflows typically impose numerous constraints on the final deliverable, including structural organization, computation logic, formatting, and executable correctness. Although these requirements are explicitly specified in the task description, current models frequently exhibit partial compliance: they satisfy the most visible requirements while overlooking other equally critical constraints. As a result, the generated artifact appears largely complete from a superficial inspection, yet fails to satisfy the complete acceptance criteria required for direct use. This behavior indicates that models often optimize for producing outputs that look correct, rather than ensuring that every instruction governing the final deliverable has been faithfully executed.

\begin{figure*}[t]
    \centering
    \includegraphics[width=0.99\textwidth]{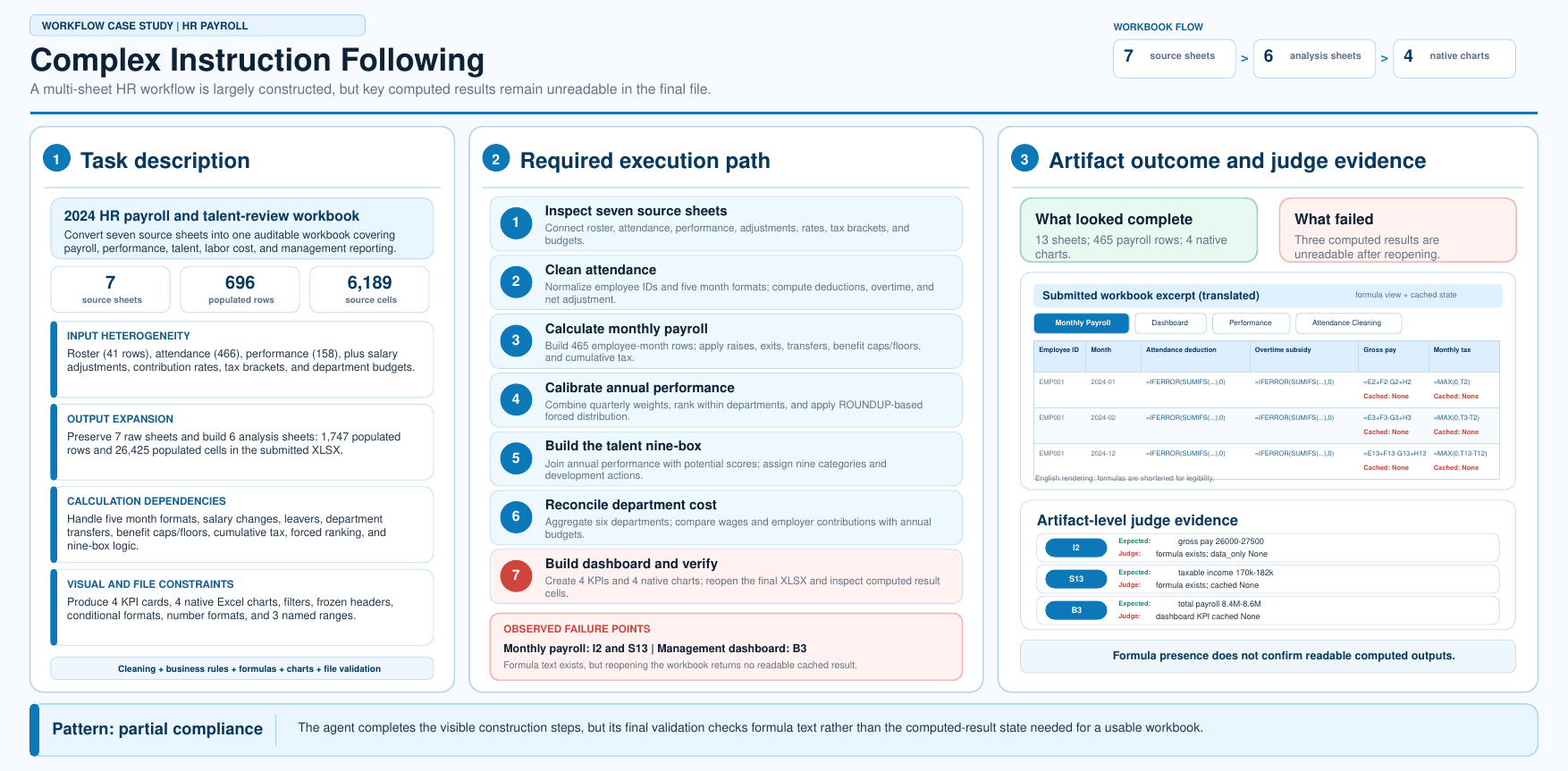} 
    \caption{
A representative case of complex instruction following failure.
Although the generated workbook appears complete, artifact-level evaluation reveals that several critical acceptance conditions remain unsatisfied, illustrating a common pattern of partial compliance.
    }
    \label{fig:instruction}
\end{figure*}

A representative example is shown in Figure~\ref{fig:instruction}. The task requires generating a complete HR payroll workbook from multiple source sheets under a diverse set of structural, computational, and formatting constraints. Although the model produces an artifact that appears largely complete—including all required worksheets, formulas, charts, and dashboard layouts—the artifact-level evaluation reveals that several critical acceptance conditions remain unsatisfied. Specifically, key cached values required for downstream verification are missing, rendering the workbook unverifiable despite its polished appearance. This case illustrates a common pattern of partial compliance, where the model satisfies the most visible requirements while overlooking less obvious but equally essential deliverable constraints.

\textbf{Self-Verification Hallucination.} We also observe that many models fail not because they cannot complete the required workflow, but because they incorrectly assume that successful execution implies successful delivery. After producing a plausible artifact, the model often summarizes the intended reasoning process or completed steps as evidence that the task has been verified, without independently inspecting the final deliverable against the original user requirements. Consequently, subtle yet critical errors—such as incorrect cell values, unmatched identifiers, missing constraints, or formatting inconsistencies—remain undetected despite an otherwise convincing output. This behavior reflects a form of self-verification hallucination, where the model mistakes confidence in its own reasoning process for evidence that the generated artifact satisfies the acceptance criteria. In real-world office workflows, however, successful completion requires validating the final deliverable itself rather than merely confirming the execution process, making this a common source of deployment-critical failures. A typical case is shown at Appendix~\ref{app:self-verify hallucination}.

\textbf{Domain-Specific Expertise.} Another major source of failure arises from insufficient domain-specific expertise, consistent with the relatively poor performance on the Domain-Specific Compliance dimension. Unlike general instruction-following errors, these failures are driven by the unique operational requirements of different professional domains. Although models often generate outputs that appear fluent and professionally formatted, they frequently fail to satisfy the domain-specific constraints that determine whether a deliverable is actually usable. As a result, the dominant failure patterns vary substantially across domains, reflecting different notions of correctness, evidence, precision, and actionability required by real-world workflows.

From a domain perspective, different professional areas expose distinct failure characteristics. Business and management tasks primarily stress artifact correctness, particularly for spreadsheets, dashboards, formulas, and structured reports. Finance tasks emphasize source attribution, numerical precision, valuation assumptions, and temporal consistency, where errors commonly arise from insufficient reconciliation, inadequate numerical self-verification, or inconsistent accounting conventions. Legal tasks are less constrained by legal writing style than by the completeness and correctness of legal reasoning, including missing citations, incomplete factual coverage, insufficient statutory support, or incorrect mapping between facts and applicable rules. Medical tasks achieve relatively higher average scores but pose substantially greater safety risks, since mistakes in medication timing, continuation criteria, or discharge planning directly undermine clinical usability. STEM and Computer Science tasks instead require precise object-level grounding, demanding accurate relationships among code, engineering drawings, images, bills of materials, and system components. Finally, education and humanities tasks place greater emphasis on fine-grained rule following, such as curriculum requirements, credit allocation, document organization, and textual coherence. A typical example of failure caused by 
insufficient domain-specific expertise is shown at Appendix~\ref{app:domain}.

\subsection{Impact of Harness on Startup Tasks}

\subsubsection{Effect of General-Purpose  Harness}
\begin{table}[t]
\centering
\small
\setlength{\tabcolsep}{7pt}
\renewcommand{\arraystretch}{1.15}
\caption{Performance under different general-purpose agent frameworks.
For each model, we report the average score under three general-purpose
frameworks and the performance range (max--min) across frameworks.}
\label{tab:harness}
\begin{tabular}{lcccc}
\toprule
\textbf{Model}
& \textbf{Hermes}
& \textbf{Claude Code}
& \textbf{Nanobot}
& \textbf{Max--Min} \\
\midrule
GPT-5.5  & 71.00 & 70.11 & 72.79 & 2.68 \\
GLM-5.1  & 60.20 & 60.18 & 60.79 & 0.61 \\
Qwen-3.6-Max & 59.10 & 57.38 & 59.46 & 2.08 \\
\midrule
Average  & 63.43 & 62.56 & 64.35 & 1.79 \\
\bottomrule
\end{tabular}
\end{table}


In the main experiment, all models are evaluated under nanobot Agent framework. To examine whether the observed performance is primarily determined by the choice of agent framework rather than the underlying model, we re-evaluate the benchmark using two more representative general-purpose agent frameworks: Hermes~\cite{hermes_agent} and Claude Code~\cite{anthropic_claude_code}. All frameworks are evaluated under the same task set and identical evaluation protocol, with only the execution framework replaced.

As shown in Table~\ref{tab:harness}, replacing the general-purpose framework leads to only minor performance differences. Across all evaluated models, the average variation between the best and worst framework is only 1.79 points. GLM-5.1 is almost completely unaffected (0.61 points), while even the largest difference observed on GPT-5.5 is only 2.68 points. More importantly, the relative ordering of models remains unchanged across all three frameworks.
These results suggest that StartupBench is largely robust to the choice of general-purpose agent framework. Although different frameworks adopt different prompting strategies, tool interfaces, and interaction mechanisms, simply replacing one general-purpose framework does not substantially alter end-to-end task performance. Consequently, the performance gap observed on StartupBench primarily reflects the capability of the underlying foundation models rather than implementation-specific characteristics of a particular framework.

\subsubsection{General-Purpose Agents versus Specialized Agents}

To further understand the role of agent frameworks, we compare general-purpose agent frameworks with the specialized startup agents from which StartupBench tasks are derived. Unlike the previous experiment, this comparison evaluates each task using its corresponding production startup agent, which has been specifically optimized for its target workflow through domain-specific model selection, prompting strategies, tool orchestration, and workflow design.

Table~\ref{tab:general_specialized} summarizes the comparison between general-purpose and specialized agent systems. Across all executions, general-purpose agents achieve an average score of 64.26 and a success rate of 19.74\%, substantially below the specialized agents at 83.50 and 39.18\%, respectively. Since specialized agents are explicitly designed and optimized for their target domains, we further consider a stronger oracle setting for the general-purpose agents: for each model--task pair, we retain the best result among three independent trials. This oracle selection raises the average score to 71.75 and success rate to 28.06\%, but still falls well short of the specialized agents, with gaps of 11.75 points in average score and 11.12 percentage points in success rate. Thus, the performance gap cannot be explained simply by stochastic variation or occasional execution failures: even when general-purpose agents are allowed multiple attempts and evaluated by their best observed outcome, specialized agent systems remain substantially more effective on these tasks.
Combined with the failure-mode analysis presented earlier, these findings provide a more complete picture of the remaining gap between general-purpose foundation models and specialized startup agents. While today's foundation models still cannot fully replace domain-specialized agents under a unified general-purpose framework, their shortcomings are concentrated in a relatively clear set of capabilities, including complex instruction following, domain-specific knowledge, professional operational conventions, and long-horizon workflow execution. This suggests that the observed advantage of specialized systems does not necessarily require fundamentally different agent architectures, but may instead reflect capabilities that current general-purpose models have yet to acquire reliably. As these underlying capabilities continue to improve, general-purpose agents may increasingly accomplish valuable professional end-to-end tasks without relying on domain-specific agent harness designs.

\begin{table}[t]
\centering
\small
\caption{General-purpose versus specialized agent systems. Oracle General denotes the averages of the highest score achieved in all trials for every model.}
\label{tab:general_specialized}
\begin{tabular}{lcc}
\toprule
\textbf{Agent System}
& \textbf{Avg Score}
& \textbf{Success Rate} \\
\midrule
General (All Runs avg) & 64.26 & 19.74 \\
General (Oracle) & 71.75 & 28.06 \\
Specialized Agent & \textbf{83.50} & \textbf{39.18} \\
\bottomrule
\end{tabular}
\end{table}

\section{Conclusion}

We introduce \textbf{StartupBench}, a benchmark for evaluating models on end-to-end professional workflows derived from market-validated AI-native products and real-world user demands. Across multiple tasks spanning diverse domains, our evaluation reveals a substantial gap between making meaningful progress on professional work and producing deliverables that fully satisfy practical acceptance criteria. This gap is primarily driven by limitations in complex instruction following, domain-specific expertise, professional conventions, and long-horizon workflow execution. By grounding evaluation in workflows that users already seek to delegate to AI, StartupBench provides a realistic measure of practical agent capabilities and highlights concrete directions for future improvement. We hope StartupBench supports the development of models and agents that can not only perform valuable professional work, but reliably deliver results ready for direct use.

\clearpage
\section{Contributions}

\textbf{Project Leads} \\
Liya Zhu, Xin Ma, Tao Liu, Haodong Wang, Ge Zhang \\

\vspace{0.5em}

\textbf{Core Contributors} \\
Jingzhe Ding, Qingshui Gu, Yongjie Zhong, Jinxiang Meng, Yuan Gao,
Yunqiu Zhou, Hao Zhu, Jifeng He,
Yongzhi Liao, Xinyi Zhang, Chaoxin Li, Yi Zhu,
Xi Lin, Duju Zeng, Xiang Gao, Wen Zhang,
Yunyang Wang, Duo Wang \\

\vspace{0.5em}

\textbf{Contributors} \\
Huan Zhou, Zuo Wang, Jin Chen, Kaiyuan Zhang, Chuqian Yu, Tianhao Yu, Longxiang Liu, Jianbo Xue, Huimin Che, Jiahao Wang \\

\vspace{0.5em}

\textbf{Sponsor Committee} \\
Yujia Qin, Jiaheng Liu

\vspace{0.5em}

\textbf{Corresponding Authors} \\
Ge Zhang (\email{gezhang@umich.edu})\\
Shen Yan  (\email{sheny@bytedance.com}) \\ 
Xiaolong Chang (\email{changxiaolong@bytedance.com})\\
Wenhao Huang (\email{huang.wenhao@bytedance.com})\\

%% file: sections/appendix.tex
\section{Tutorial of StartupBench Task Annotation}

\subsection{Task Construction}
\label{app:task_construction}

A \textsc{startup} task can be defined as a triple $\mathcal{T}=(q,\mathcal{E},\mathcal{R})$.
The experts are required to provide all the elements of a task:

\begin{itemize}
    \item \textbf{Input query}: The query $q$ is a natural-language statement of the task, specifying the user instruction that initiates the task. Derived from authentic scenarios of the work of annotators, the input query is supposed to reflect realistic user intent, including task background, implicit constraints and expected deliverables.

    \item \textbf{Task workspace}: The environment $\mathcal{E}$ denotes the complete workspace provided to the agent, including the multimodal source files together with any additional resources and interaction setting required to complete the task. The task workspace provides a self-contained execution environment for the agent. Each workspace includes all necessary input artifacts (e.g., files, datasets, or structured resources) required to complete the task, and ensures that the agent has access to sufficient input information to solve the problem. This design enables reproducible evaluation under controlled conditions while preserving realistic workflow structure.

    \item \textbf{Evaluation rubrics}: The rubric $\mathcal{R}=\{(p_i,w_i)\}_{i=1}^{n}$ is a checklist within which every item consists of a natural-language scoring point $p_i$ and a positive importance weight $w_i$ reflecting its importance to successful task completion. The evaluation rubrics define a set of fine-grained criteria used to assess task completion quality. Due to the complexity and multi-faceted nature of real-world workflows, each task is evaluated using multiple rubric items covering different dimensions. We categorize all rubrics in StartupBench into 6 dimensions and 3 importance levels, which are detailed in Appendix~\ref{app:rubrics}.
    
\end{itemize}

Together, these components preserve the structure of real-world agent workflows while enabling systematic and objective evaluation. Tasks may require information synthesis, constraint following, structured generation, or domain-specific reasoning and judgment.

\subsection{Cross Validation}
\label{app:cross_valid}

To ensure reliability of the quality of data, in our construction process each task is independently reviewed by at least one additional domain expert with relevant professional experience. Rather than focusing only on annotation consistency, reviewers validate the realism and correctness of the task from multiple complementary perspectives. 

\begin{itemize} \item \textbf{Task authenticity.} Reviewers verify the authenticity of both the task description and the accompanying workspace. This includes checking that the input query reflects realistic user requests, that all provided materials are factually correct, and that no domain-specific knowledge errors exist. 

\item \textbf{Workflow fidelity.} Reviewers examine whether the task faithfully represents real-world workflows in the corresponding profession. Tasks that deviate from practical working procedures or fail to reflect genuine job responsibilities are revised or discarded.

\item \textbf{Rubric validity.} Reviewers inspect the evaluation rubrics to ensure that the assessed capabilities are meaningful for the target workflow, that rubric items are clear and unambiguous, and that their relative weights reasonably reflect task priorities. 

\item \textbf{Ground-truth verification.} Reviewers validate the reference solution (ground truth) by evaluating it against the finalized rubrics. The ground-truth deliverable is required to achieve a full score. If inconsistencies are identified between the reference solution and the evaluation criteria, both the solution and the rubrics are revised until they are mutually consistent. \end{itemize}

\section{Domain Expert Background}
\label{app:expert background}
We recruit 57 domain experts to support task construction and cross-validation. The expert pool covers all 6 domains in StartupBench and spans a diverse range of professional backgrounds, including clinical medicine, law, investment and quantitative finance, software development and artificial intelligence, engineering, business management and consulting, and education and humanities. Experts are assigned to tasks according to their reported areas of expertise and professional experience, ensuring that task construction and review were conducted by individuals familiar with the corresponding professional domains.

Table~\ref{tab:expert_background} summarizes the professional experience of the expert pool. The median professional experience is 5 years. Among the 57 experts, 68.4\% have at least 3 years of professional experience, 42.1\% have at least 6 years, and 24.6\% have at least 10 years. These backgrounds provide practical domain knowledge for translating interview-derived user scenarios into reproducible benchmark tasks and for assessing the realism, correctness, and evaluation criteria of tasks during cross-validation. All experts are reasonably compensated based on the actual workload associated with task construction and review.

\begin{table}[t]
\centering
\small
\caption{Summary of domain expert backgrounds. Professional experience refers to reported years of experience in the corresponding or closely related domain.}
\label{tab:expert_background}
\begin{tabular}{lc}
\toprule
\textbf{Expert Statistics} & \textbf{Value} \\
\midrule
Number of domain experts                  & 57    \\
Median professional experience            & 5 years \\
Experts with $\geq 3$ years of experience & 68.4\% \\
Experts with $\geq 6$ years of experience & 42.1\% \\
Experts with $\geq 10$ years of experience & 24.6\% \\
StartupBench domain coverage              & 6 / 6 \\
\bottomrule
\end{tabular}
\end{table}

\section{Rubrics of StartupBench}
\label{app:rubrics}

\textbf{Rubric Dimensions.} To facilitate consistent evaluation across heterogeneous real-world tasks, we organize all evaluation rubrics into six high-level categories according to the primary capability they assess. Rather than being tied to specific application domains, these categories capture common dimensions of deliverable quality that recur across office tasks, including correctness, completeness, data processing, professional reasoning, engineering quality, and presentation. This taxonomy enables more interpretable analysis of model strengths and weaknesses while maintaining a unified evaluation framework across different task types.

\begin{itemize}
    \item \textbf{Structure \& Completeness.}
    Evaluates whether the required deliverables are fully produced with the correct organization, file structure, required components, and overall completeness.

    \item \textbf{Calculation Precision.}
    Evaluates the correctness of numerical results, logical reasoning, formula execution, factual consistency, and other objective computations.

    \item \textbf{Information Integration.}
    Evaluates data cleaning, transformation, aggregation, statistical analysis, feature engineering, modeling, and other structured data manipulation workflows.

    \item \textbf{Domain-Specific Compliance.}
    Evaluates whether outputs satisfy domain-specific professional requirements, including business logic, financial principles, legal reasoning, medical practice, educational standards, and other expert knowledge.

    \item \textbf{Engineering \& Format.}
    Evaluates implementation quality, coding conventions, spreadsheet engineering, document formatting, reproducibility, robustness, naming conventions, and compliance with technical specifications.

    \item \textbf{Output \& Presentation.}
    Evaluates the quality of visual presentation and communication, including charts, layouts, formatting, readability, report organization, and overall usability of the final deliverables.
\end{itemize}

Table~\ref{tab:rubric_distribution} summarizes the distribution of all 2,453 rubrics in StartupBench.  Calculation Precision constitutes the largest rubric category (38.9\%), reflecting the importance of producing objectively correct work products. Structure \& Completeness and Domain-Specific Compliance together account for another 42.8\%, highlighting that successful completion of realistic office workflows requires not only correct computations but also complete deliverables and professional domain reasoning.

\begin{table}[h]
\centering
\small
\caption{Distribution of rubric categories in StartupBench.}
\label{tab:rubric_distribution}
\begin{tabular}{lc}
\toprule
\textbf{Category} & \textbf{Percentage} \\
\midrule
Structure \& Completeness & 27.31\% \\
Calculation Precision & 38.89\% \\
Information Integration & 9.46\% \\
Domain-Specific Compliance & 15.49\% \\
Engineering \& Format & 6.20\% \\
Output \& Presentation & 2.65\% \\
\midrule
\textbf{Total} & \textbf{100.00\%} \\
\bottomrule
\end{tabular}
\end{table}

\begin{figure*}[t]
    \centering
    \includegraphics[width=0.99\textwidth]{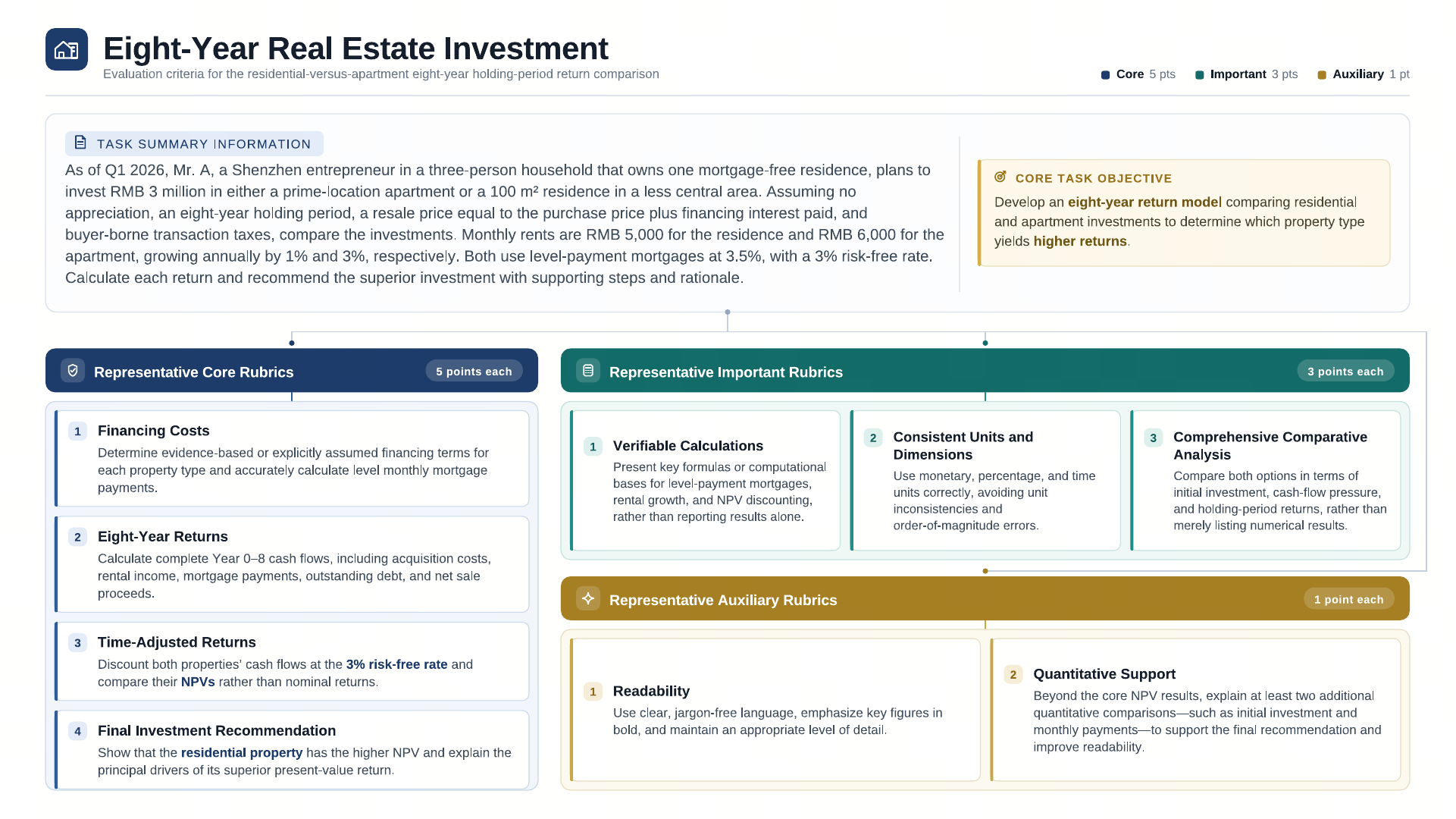}
    \caption{
A representative example of a rubric of a task spanning 3 levels. Core rubrics assess the essential financial reasoning and calculations required to solve the investment problem, Important rubrics evaluate the correctness and completeness of the supporting analysis, while Auxiliary rubrics capture presentation quality and additional quantitative support.}
    \label{fig:rubric_level}
\end{figure*}

\textbf{Rubric Importance Levels.} To account for differences in the importance of individual requirements to successful task completion, we assign each rubric to one of three importance levels. \textbf{Core} rubrics capture requirements that directly determine whether the primary user needs are satisfied; violating them would substantially compromise the correctness, reliability, safety, or usability of the final deliverable. \textbf{Important} rubrics represent key requirements for high-quality task completion whose satisfaction materially affects the overall quality of the deliverable, but whose individual violation does not necessarily render it unusable. \textbf{Auxiliary} rubrics capture lower-level requirements whose omission primarily affects details, user experience, visual appeal, or polish rather than the fundamental usability of the deliverable. We assign weights of \textbf{5}, \textbf{3}, and \textbf{1} to Core, Important, and Auxiliary rubrics, respectively, and use these weights when aggregating rubric-level judgments into the task score. A typical case of task containing rubrics of different levels is shown at Figure~\ref{fig:rubric_level}.

Across  all tasks, Core, Important, and Auxiliary rubrics account for an average of 59.10\%, 34.33\%, and 6.57\% of the total task weight, respectively, ensuring that task scores are primarily determined by requirements that materially affect successful task completion.

\section{Examples of Behavior-level Failure Modes}

\subsection{Self-Verification Hallucination}
\label{app:self-verify hallucination}

\begin{figure*}[t]
    \centering
    \includegraphics[width=0.99\textwidth]{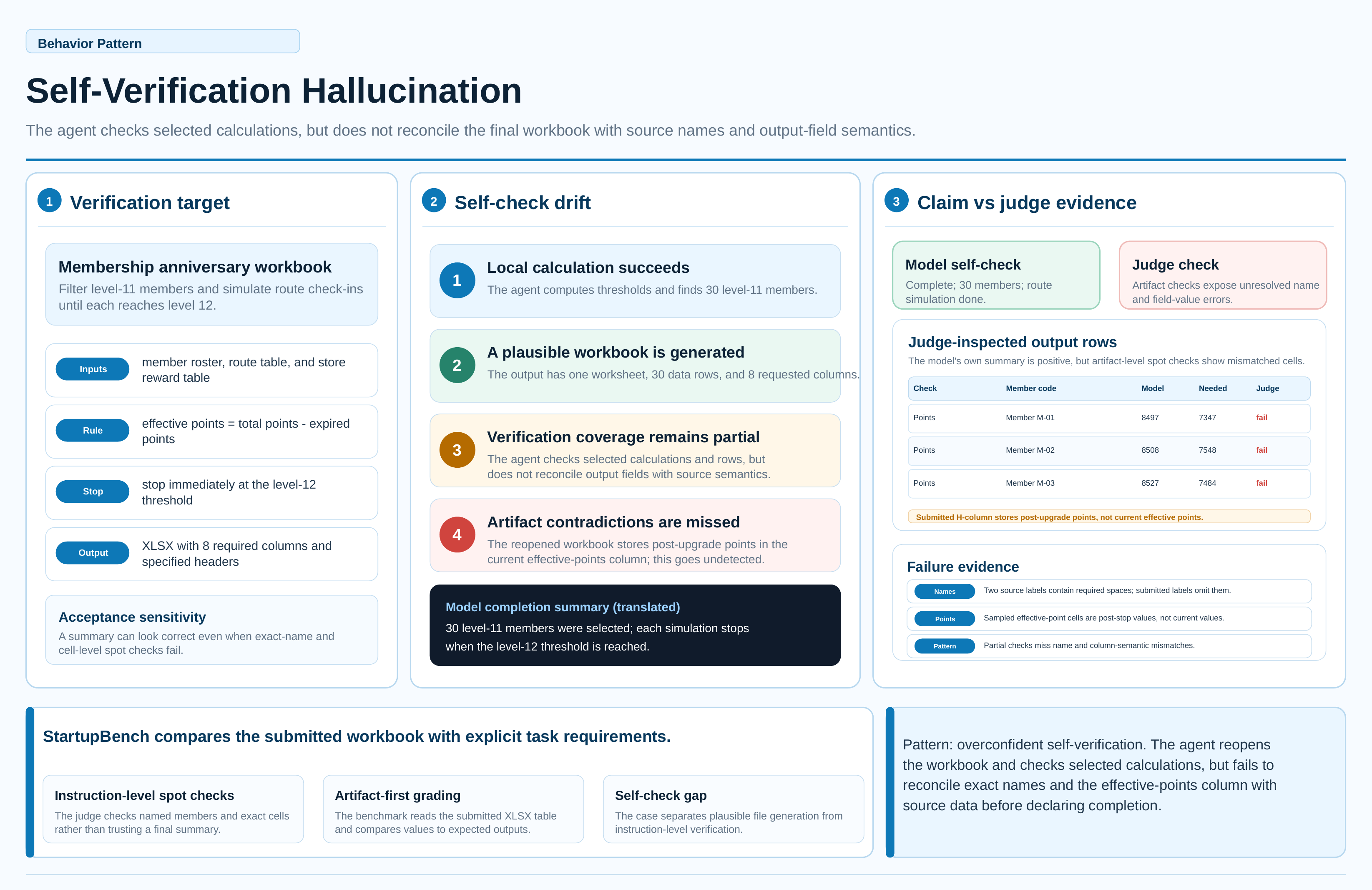}
    \caption{
A representative example of self-verification hallucination. The model generates a plausible workbook and concludes that the task has been completed based on its execution process. However, artifact-level spot checks expose acceptance-critical errors in the final deliverable, showing that the model mistakes a progress summary for actual verification instead of validating the generated artifact against the original task requirements.}
    \label{fig:self_verify}
\end{figure*}

A representative example is shown in Fig.~\ref{fig:self_verify}. The task requires the model to filter all level-11 members, simulate route check-ins until each member reaches the level-12 threshold, and generate an Excel workbook satisfying a set of precise value and formatting constraints. The produced workbook appears largely complete, containing the required worksheet, member records, and output columns, leading the model to conclude that the task has been successfully finished. However, artifact-level inspection reveals multiple acceptance-critical errors, including incorrect effective-point values and mismatched member names. The underlying issue is not that the model fails to perform the required computation, but that it never verifies the generated artifact itself. Instead, it treats a successful execution summary as evidence that the final deliverable has been validated. Consequently, localized yet user-critical errors remain unnoticed, illustrating a form of \emph{self-verification hallucination}, where confidence is derived from the reasoning process rather than from independently checking the completed artifact against the original task requirements.

\subsection{Domain-Specific Failure}
\label{app:domain}

\begin{figure*}[t]
    \centering
    \includegraphics[width=0.99\textwidth]{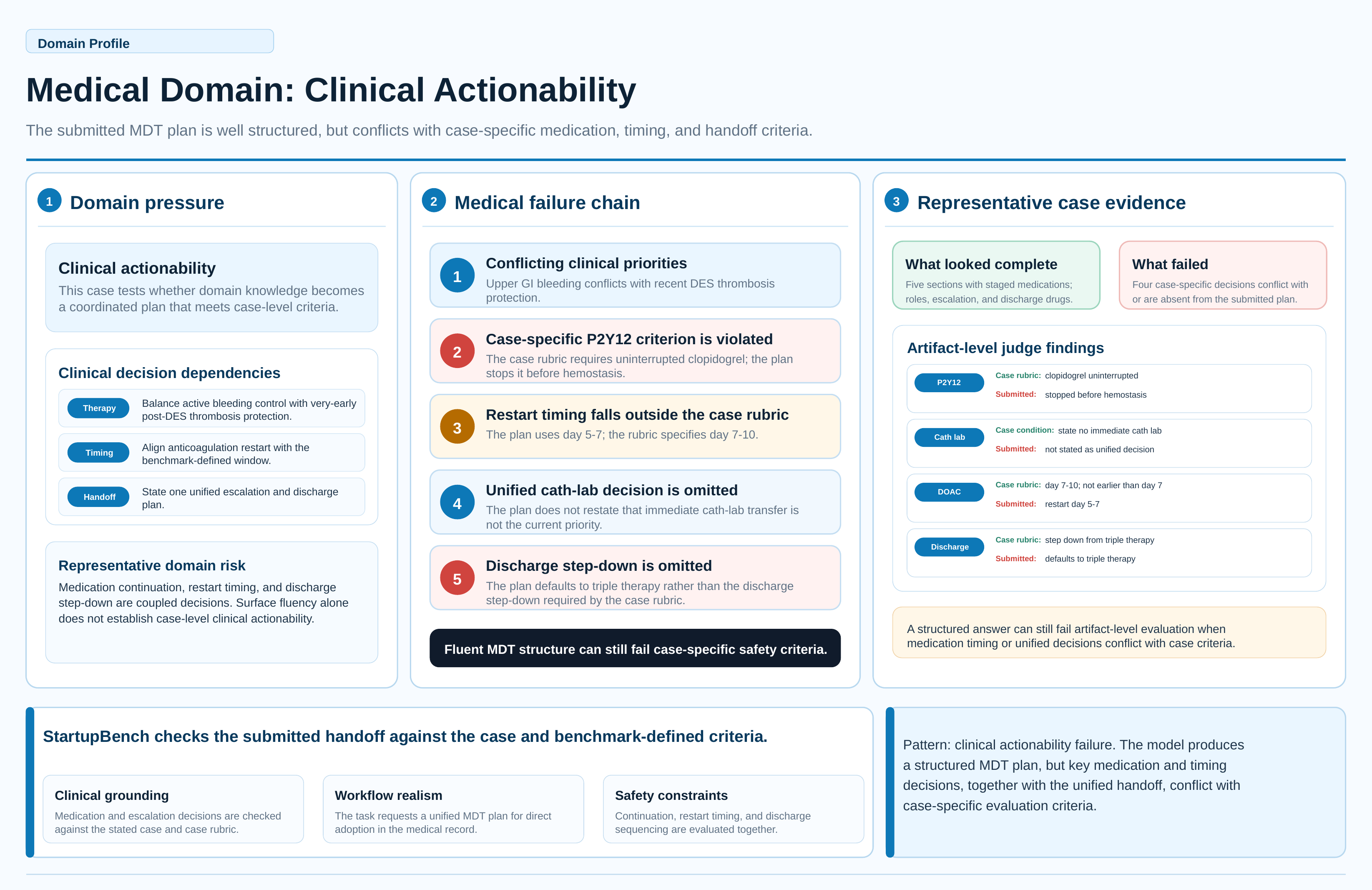}
    \caption{
A representative example of failure caused by insufficient domain-specific clinical actionability.}
    \label{fig:domain-specify}
\end{figure*}

A representative example of failure caused by insufficient domain-specific clinical actionability is shown at Figure~\ref{fig:domain-specify}. The model generates a well-formatted multidisciplinary treatment plan that appears clinically professional, yet violates multiple safety-critical treatment constraints. The failure stems not from poor writing quality, but from an inability to convert medical knowledge into actionable and clinically reliable decisions.

%% file: paper.bbl
\begin{thebibliography}{36}
\providecommand{\natexlab}[1]{#1}
\providecommand{\url}[1]{\texttt{#1}}
\expandafter\ifx\csname urlstyle\endcsname\relax
  \providecommand{\doi}[1]{doi: #1}\else
  \providecommand{\doi}{doi: \begingroup \urlstyle{rm}\Url}\fi

\bibitem[{Anthropic}()]{anthropic_claude_code}
{Anthropic}.
\newblock Claude code.
\newblock \url{https://github.com/anthropics/claude-code}.
\newblock GitHub repository.

\bibitem[{ByteDance Seed}(2026)]{bytedance2026seed21}
{ByteDance Seed}.
\newblock Seed2.1 model card: Agentic intelligence for productivity.
\newblock Technical report, ByteDance, 2026.
\newblock URL \url{https://lf3-static.bytednsdoc.com/obj/eden-cn/lapzild-tss/ljhwZthlaukjlkulzlp/seed2.1/Seed2_1_Model_Card.pdf}.

\bibitem[DeepSeek-AI(2026)]{deepseek2026v4}
DeepSeek-AI.
\newblock Deepseek-v4: Towards highly efficient million-token context intelligence, 2026.
\newblock URL \url{https://huggingface.co/collections/deepseek-ai/deepseek-v4}.
\newblock Technical Report.

\bibitem[Drouin et~al.()Drouin, Gasse, Caccia, Laradji, Del~Verme, Marty, Boisvert, Thakkar, Cappart, Vazquez, et~al.]{drouin2403workarena}
Alexandre Drouin, Maxime Gasse, Massimo Caccia, Issam~H Laradji, Manuel Del~Verme, Tom Marty, L{\'e}o Boisvert, Megh Thakkar, Quentin Cappart, David Vazquez, et~al.
\newblock Workarena: How capable are web agents at solving common knowledge work tasks?, 2024.
\newblock \emph{URL https://arxiv. org/abs/2403.07718}.

\bibitem[Du et~al.(2026)Du, Yao, Ma, Wang, Zheng, Liu, Liang, Jin, Wei, Zheng, et~al.]{du2026supergpqa}
Xeron Du, Yifan Yao, Kaijing Ma, Bingli Wang, Tianyu Zheng, Minghao Liu, Yiming Liang, Xiaolong Jin, Zhenlin Wei, Chujie Zheng, et~al.
\newblock Supergpqa: Scaling llm evaluation across 285 graduate disciplines.
\newblock \emph{Advances in Neural Information Processing Systems}, 38, 2026.

\bibitem[{Google Cloud}(2026)]{GoogleCloud2026Gemini3.1Pro}
{Google Cloud}.
\newblock {Gemini 3.1 Pro Model Card}, 4 2026.
\newblock URL \url{https://docs.cloud.google.com/vertex-ai/generative-ai/docs/models/gemini/3-1-pro?hl=zh-cn}.

\bibitem[Hu et~al.(2026)Hu, Zhang, Huang, Tu, Su, Deng, Liu, Liu, Liu, and Ho]{hu2026occubenchevaluatingaiagents}
Xiaomeng Hu, Yinger Zhang, Fei Huang, Jianhong Tu, Yang Su, Lianghao Deng, Yuxuan Liu, Yantao Liu, Dayiheng Liu, and Tsung-Yi Ho.
\newblock Occubench: Evaluating ai agents on real-world professional tasks via language environment simulation, 2026.
\newblock URL \url{https://arxiv.org/abs/2604.10866}.

\bibitem[Kim et~al.(2024)Kim, Shin, Jang, Longpre, Lee, Yun, Shin, Kim, Thorne, Seo, et~al.]{kim2024prometheus}
Seungone Kim, Jay Shin, Joel Jang, Shayne Longpre, Hwaran Lee, Sangdoo Yun, Ryan Shin, Sungdong Kim, James Thorne, Minjoon Seo, et~al.
\newblock Prometheus: Inducing fine-grained evaluation capability in language models.
\newblock In \emph{International Conference on Learning Representations}, 2024.

\bibitem[Lei et~al.(2025)Lei, Meng, Huang, Zhao, Zhang, Luo, Zou, Yang, Shi, Gao, et~al.]{lei2025dacomp}
Fangyu Lei, Jinxiang Meng, Yiming Huang, Junjie Zhao, Yitong Zhang, Jianwen Luo, Xin Zou, Ruiyi Yang, Wenbo Shi, Yan Gao, et~al.
\newblock Dacomp: Benchmarking data agents across the full data intelligence lifecycle.
\newblock \emph{arXiv preprint arXiv:2512.04324}, 2025.

\bibitem[Liu et~al.(2024)Liu, Yu, Zhang, Xu, Lei, Lai, Gu, Ding, Men, Yang, et~al.]{liu2024agentbench}
Xiao Liu, Hao Yu, Hanchen Zhang, Yifan Xu, Xuanyu Lei, Hanyu Lai, Yu~Gu, Hangliang Ding, Kaiwen Men, Kejuan Yang, et~al.
\newblock Agentbench: Evaluating llms as agents.
\newblock In \emph{International Conference on Learning Representations}, 2024.

\bibitem[Meng et~al.(2026)Meng, Huang, Lei, Guo, Liu, Su, Wang, Wang, Wang, Yang, et~al.]{meng2026dvworld}
Jinxiang Meng, Shaoping Huang, Fangyu Lei, Jingyu Guo, Haoxiang Liu, Jiahao Su, Sihan Wang, Yao Wang, Enrui Wang, Ye~Yang, et~al.
\newblock Dv-world: Benchmarking data visualization agents in real-world scenarios.
\newblock \emph{arXiv preprint arXiv:2604.25914}, 2026.

\bibitem[Mialon et~al.(2024)Mialon, Fourrier, Wolf, LeCun, and Scialom]{mialon2024gaia}
Gr{\'e}goire Mialon, Cl{\'e}mentine Fourrier, Thomas Wolf, Yann LeCun, and Thomas Scialom.
\newblock Gaia: a benchmark for general ai assistants.
\newblock In \emph{International Conference on Learning Representations}, 2024.

\bibitem[{Moonshot AI}(2026)]{moonshot2026k2_6}
{Moonshot AI}.
\newblock {Kimi K2.6}: Advancing open-source coding.
\newblock \url{https://www.kimi.com/blog/kimi-k2-6}, 2026.
\newblock Accessed: 2026-06-02.

\bibitem[{Nous Research}(2026)]{hermes_agent}
{Nous Research}.
\newblock Hermes agent: The ai agent that learns from you.
\newblock \url{https://github.com/hermes-agent-org/hermes}, 2026.
\newblock GitHub repository.

\bibitem[{OpenAI}(2026)]{openai2026gpt56}
{OpenAI}.
\newblock Gpt-5.6 system card.
\newblock Technical report, OpenAI, 7 2026.
\newblock URL \url{https://deploymentsafety.openai.com/gpt-5-6/gpt-5-6.pdf}.

\bibitem[OpenAI(2026)]{openai_gpt55}
OpenAI.
\newblock Introducing gpt-5.5, 4 2026.
\newblock URL \url{https://openai.com/index/introducing-gpt-5-5/}.
\newblock System Card: https://deploymentsafety.openai.com/gpt-5-5/gpt-5-5.pdf.

\bibitem[Opsahl-Ong et~al.(2026)Opsahl-Ong, Singhvi, Collins, Zhou, Wang, Baheti, Oertell, Portes, Havens, Elsen, et~al.]{opsahl2026officeqa}
Krista Opsahl-Ong, Arnav Singhvi, Jasmine Collins, Ivan Zhou, Cindy Wang, Ashutosh Baheti, Owen Oertell, Jacob Portes, Sam Havens, Erich Elsen, et~al.
\newblock Officeqa pro: An enterprise benchmark for end-to-end grounded reasoning.
\newblock \emph{arXiv preprint arXiv:2603.08655}, 2026.

\bibitem[Patwardhan et~al.(2025)Patwardhan, Dias, Proehl, Kim, Wang, Watkins, Fishman, Aljubeh, Thacker, Fauconnet, et~al.]{patwardhan2025gdpval}
Tejal Patwardhan, Rachel Dias, Elizabeth Proehl, Grace Kim, Michele Wang, Olivia Watkins, Sim{\'o}n~Posada Fishman, Marwan Aljubeh, Phoebe Thacker, Laurance Fauconnet, et~al.
\newblock Gdpval: Evaluating ai model performance on real-world economically valuable tasks.
\newblock \emph{arXiv preprint arXiv:2510.04374}, 2025.

\bibitem[Phan et~al.(2025)Phan, Gatti, Han, Li, Hu, Zhang, Zhang, Shaaban, Ling, Shi, et~al.]{phan2025humanity}
Long Phan, Alice Gatti, Ziwen Han, Nathaniel Li, Josephina Hu, Hugh Zhang, Chen Bo~Calvin Zhang, Mohamed Shaaban, John Ling, Sean Shi, et~al.
\newblock Humanity's last exam.
\newblock \emph{arXiv preprint arXiv:2501.14249}, 2025.

\bibitem[{Qwen Team}(2026)]{qwen36_max_preview}
{Qwen Team}.
\newblock {Qwen3.6-Max-Preview}: Smarter, sharper, still evolving, April 2026.
\newblock URL \url{https://qwen.ai/blog?id=qwen3.6-max-preview}.

\bibitem[Schick et~al.(2023)Schick, Dwivedi-Yu, Dess{\`\i}, Raileanu, Lomeli, Hambro, Zettlemoyer, Cancedda, and Scialom]{schick2023toolformer}
Timo Schick, Jane Dwivedi-Yu, Roberto Dess{\`\i}, Roberta Raileanu, Maria Lomeli, Eric Hambro, Luke Zettlemoyer, Nicola Cancedda, and Thomas Scialom.
\newblock Toolformer: Language models can teach themselves to use tools.
\newblock \emph{Advances in neural information processing systems}, 36:\penalty0 68539--68551, 2023.

\bibitem[Shi et~al.(2026)Shi, Wang, Zhao, Chen, Feng, Hao, Su, Gu, Su, Cai, et~al.]{shi2026aj}
Wentao Shi, Yu~Wang, Yuyang Zhao, Yuxin Chen, Fuli Feng, Xueyuan Hao, Xi~Su, Qi~Gu, Hui Su, Xunliang Cai, et~al.
\newblock Aj-bench: Benchmarking agent-as-a-judge for environment-aware evaluation.
\newblock \emph{arXiv preprint arXiv:2604.18240}, 2026.

\bibitem[Sun et~al.(2026)Sun, Han, Zhang, Pang, Wang, Cao, Huang, Duroiu, Zhang, Lin, et~al.]{sun2026agents}
Yiyou Sun, Xinyang Han, Weichen Zhang, Yuanbo Pang, Tianyu Wang, Yuhan Cao, Yixiao Huang, Chris Duroiu, Haoyun Zhang, Jeffrey Lin, et~al.
\newblock Agents' last exam.
\newblock \emph{arXiv preprint arXiv:2606.05405}, 2026.

\bibitem[Tang et~al.(2026)Tang, Zhou, Liu, Li, Wu, Wang, Huang, Zhou, Zhou, Song, et~al.]{tang2026workspace}
Zirui Tang, Xuanhe Zhou, Yumou Liu, Linchun Li, Yukai Wu, Weizheng Wang, Hongzhang Huang, Wei Zhou, Jun Zhou, Jiachen Song, et~al.
\newblock Workspace-bench 1.0: Benchmarking ai agents on workspace tasks with large-scale file dependencies.
\newblock \emph{arXiv preprint arXiv:2605.03596}, 2026.

\bibitem[Team et~al.(2026)Team, Bai, Bai, Bao, Cai, Cai, Cao, Cao, Chai, Charles, et~al.]{team2026kimi}
Kimi Team, Tongtong Bai, Yifan Bai, Yiping Bao, Jianfeng Cai, Xinyuan Cai, Peizhou Cao, Yuxuan Cao, Ziwei Chai, Y~Charles, et~al.
\newblock Kimi k3: Open frontier intelligence.
\newblock \emph{arXiv preprint arXiv:2607.24653}, 2026.

\bibitem[Wang et~al.(2024)Wang, Ma, Zhang, Ni, Chandra, Guo, Ren, Arulraj, He, Jiang, et~al.]{wang2024mmlu}
Yubo Wang, Xueguang Ma, Ge~Zhang, Yuansheng Ni, Abhranil Chandra, Shiguang Guo, Weiming Ren, Aaran Arulraj, Xuan He, Ziyan Jiang, et~al.
\newblock Mmlu-pro: A more robust and challenging multi-task language understanding benchmark.
\newblock \emph{Advances in Neural Information Processing Systems}, 37:\penalty0 95266--95290, 2024.

\bibitem[Xie et~al.(2024)Xie, Zhang, Chen, Li, Zhao, Cao, Hua, Cheng, Shin, Lei, Liu, Xu, Zhou, Savarese, Xiong, Zhong, and Yu]{xieosworld}
Tianbao Xie, Danyang Zhang, Jixuan Chen, Xiaochuan Li, Siheng Zhao, Ruisheng Cao, Toh~Jing Hua, Zhoujun Cheng, Dongchan Shin, Fangyu Lei, Yitao Liu, Yiheng Xu, Shuyan Zhou, Silvio Savarese, Caiming Xiong, Victor Zhong, and Tao Yu.
\newblock {OSW}orld: Benchmarking multimodal agents for open-ended tasks in real computer environments.
\newblock In \emph{The Thirty-eight Conference on Neural Information Processing Systems Datasets and Benchmarks Track}, 2024.

\bibitem[Xu et~al.(2026)Xu, Song, Li, Tang, Jain, Bao, Wang, Zhou, Guo, Cao, et~al.]{xu2026theagentcompany}
Frank~Fangzheng Xu, Yufan Song, Boxuan Li, Yuxuan Tang, Kritanjali Jain, Mengxue Bao, Zora Wang, Xuhui Zhou, Zhitong Guo, Murong Cao, et~al.
\newblock Theagentcompany: benchmarking llm agents on consequential real world tasks.
\newblock \emph{Advances in Neural Information Processing Systems}, 38, 2026.

\bibitem[Yang et~al.(2026)Yang, Liu, Li, Bai, Chen, Chen, Duan, Dong, Hu, Jia, et~al.]{yang2026onemillion}
Qianyu Yang, Yang Liu, Jiaqi Li, Jun Bai, Hao Chen, Kaiyuan Chen, Tiliang Duan, Jiayun Dong, Xiaobo Hu, Zixia Jia, et~al.
\newblock Onemillion-bench: How far are language agents from human experts?
\newblock \emph{arXiv preprint arXiv:2603.07980}, 2026.

\bibitem[Yao et~al.(2023)Yao, Zhao, Yu, Du, Shafran, Narasimhan, and Cao]{yao2023react}
Shunyu Yao, Jeffrey Zhao, Dian Yu, Nan Du, Izhak Shafran, Karthik Narasimhan, and Yuan Cao.
\newblock React: Synergizing reasoning and acting in language models.
\newblock In \emph{11th International Conference on Learning Representations, ICLR 2023}, 2023.

\bibitem[You et~al.(2026)You, Cai, Zhang, Xu, Liu, Yu, Li, and Li]{you2026agent}
Runyang You, Hongru Cai, Caiqi Zhang, Qiancheng Xu, Meng Liu, Tiezheng Yu, Yongqi Li, and Wenjie Li.
\newblock Agent-as-a-judge.
\newblock \emph{arXiv preprint arXiv:2601.05111}, 2026.

\bibitem[Zeng et~al.(2026)Zeng, Lv, Hou, Du, Zheng, Chen, Yin, Ge, Huang, Xie, et~al.]{zeng2026glm}
Aohan Zeng, Xin Lv, Zhenyu Hou, Zhengxiao Du, Qinkai Zheng, Bin Chen, Da~Yin, Chendi Ge, Chenghua Huang, Chengxing Xie, et~al.
\newblock Glm-5: from vibe coding to agentic engineering.
\newblock \emph{arXiv preprint arXiv:2602.15763}, 2026.

\bibitem[Zheng et~al.(2023)Zheng, Chiang, Sheng, Zhuang, Wu, Zhuang, Lin, Li, Li, Xing, et~al.]{zheng2023judging}
Lianmin Zheng, Wei-Lin Chiang, Ying Sheng, Siyuan Zhuang, Zhanghao Wu, Yonghao Zhuang, Zi~Lin, Zhuohan Li, Dacheng Li, Eric Xing, et~al.
\newblock Judging llm-as-a-judge with mt-bench and chatbot arena.
\newblock \emph{Advances in neural information processing systems}, 36:\penalty0 46595--46623, 2023.

\bibitem[Zhou et~al.(2024)Zhou, Xu, Zhu, Zhou, Lo, Sridhar, Cheng, Ou, Bisk, Fried, et~al.]{zhou2024webarena}
Shuyan Zhou, Frank~F Xu, Hao Zhu, Xuhui Zhou, Robert Lo, Abishek Sridhar, Xianyi Cheng, Tianyue Ou, Yonatan Bisk, Daniel Fried, et~al.
\newblock Webarena: A realistic web environment for building autonomous agents.
\newblock In \emph{International Conference on Learning Representations}, 2024.

\bibitem[Zhu et~al.(2026)Zhu, Ding, Zhang, Xue, Liang, Zhang, Zhu, Zeng, Gao, Gu, Gao, Che, Zhao, Zhou, Wang, Xian, Le, Wu, Liu, Long, Yang, Xu, Wu, Duan, He, Li, Wang, Zhou, Hou, Yu, Shi, Gao, Chen, Chen, Luo, Zhang, Yao, Hua, Jiang, Chen, Chen, Hu, Li, Jiang, Cao, Long, Wang, Wang, Zhang, Dai, Zhang, Wang, Liu, Liu, Zhang, Chen, Qin, Zhou, Wu, Liu, Liu, Zhang, Yan, Huang, Wang, and Chang]{zhu2026workflowgymlonghorizonevaluationcomputeruse}
Liya Zhu, Jingzhe Ding, Jian Zhang, Jianbo Xue, Shihao Liang, Ge~Zhang, Yi~Zhu, Duju Zeng, Xiang Gao, Qingshui Gu, Mailun Gao, Huimin Che, Yan Zhao, Peiheng Zhou, Haojun Wang, Chaobo Xian, Lili Le, Chi Wu, Yiwei Liu, Shengda Long, Jiale Yang, Fangzhi Xu, Sijin Wu, Haodong Duan, Chao He, Zhaojian Li, Minchao Wang, Huan Zhou, Jiani Hou, Chuqian Yu, Weiran Shi, Hongwan Gao, Jiamin Chen, Guanhong Chen, Tingqin Luo, Kaiyuan Zhang, Zhixin Yao, Qing Hua, Yuhao Jiang, Jin Chen, Pu~Chen, Zhenyu Hu, Xingyu Li, Zhengxuan Jiang, Meng Cao, Tianfeng Long, Haozhe Wang, Mingzhang Wang, Yichen Zhang, Yiming Dai, Chenchen Zhang, Jiaying Wang, Xinying Liu, Xingzu Liu, Lingling Zhang, Xinjie Chen, Yujia Qin, Wangchunshu Zhou, Zhiyong Wu, Yang Liu, Jiaheng Liu, Lei Zhang, Shen Yan, Wenhao Huang, Zaiyuan Wang, and Xiaolong Chang.
\newblock Workflow-gym: Towards long-horizon evaluation of computer-use agentic tasks in real-world professional fields, 2026.
\newblock URL \url{https://arxiv.org/abs/2606.11042}.

\bibitem[Zhuge et~al.(2025)Zhuge, Zhao, Ashley, Wang, Khizbullin, Xiong, Liu, Chang, Krishnamoorthi, Tian, Shi, Chandra, and Schmidhuber]{zhuge2024agent}
Mingchen Zhuge, Changsheng Zhao, Dylan~R. Ashley, Wenyi Wang, Dmitrii Khizbullin, Yunyang Xiong, Zechun Liu, Ernie Chang, Raghuraman Krishnamoorthi, Yuandong Tian, Yangyang Shi, Vikas Chandra, and J{\"u}rgen Schmidhuber.
\newblock Agent-as-a-judge: Evaluate agents with agents.
\newblock In \emph{Forty-second International Conference on Machine Learning}, 2025.

\end{thebibliography}
